\documentclass[11pt]{article}

\usepackage[preprint]{acl}

\usepackage{times}
\usepackage{latexsym}

\usepackage[T1]{fontenc}
\usepackage{amsfonts}
\usepackage{amsmath}

\usepackage[utf8]{inputenc}
\usepackage{enumitem}
\usepackage{algorithm}
\usepackage{algorithmic}

\usepackage{microtype}

\usepackage{inconsolata}

\usepackage{graphicx}
\usepackage{amsmath}

\usepackage{multirow}
\usepackage{booktabs}
\usepackage[table]{xcolor}

\definecolor{best}{RGB}{201, 218, 248}     
\definecolor{second}{RGB}{232, 239, 251}   
\usepackage{pifont}

\usepackage{eso-pic}       

\usepackage{tcolorbox}
\tcbuselibrary{breakable}  
\newtcolorbox{promptbox}[1][]{                                      
colback=gray!3, colframe=black!60,                            
fonttitle=\bfseries\small, title={#1},                              
breakable, left=5pt, right=5pt, top=3pt, bottom=3pt,                
fontupper=\small\ttfamily,
}      

\newcommand{\ours}{Skill0.5}

\AddToShipoutPictureBG{%
  \ifnum\value{page}=1
    \AtPageUpperLeft{%
      \put(\LenToUnit{2.5cm},\LenToUnit{-2.5cm}){%
        \includegraphics[height=0.8cm]{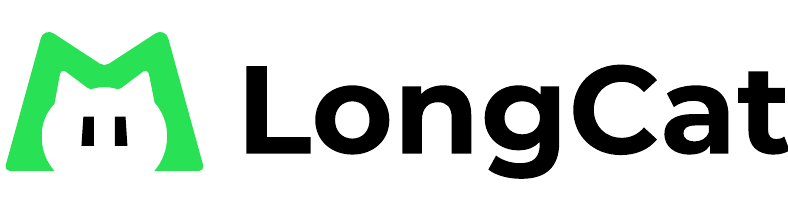}%
      }%
    }%
  \fi
}

\title{Skill0.5: Joint Skill Internalization and Utilization for Out-of-Distribution Generalization in Agentic Reinforcement Learning}

\author{
 \textbf{Jiapeng Zhu\textsuperscript{1,2,$\ast$}},
 \textbf{Jianxiang Yu\textsuperscript{1}},
 \textbf{Yibo Zhao\textsuperscript{1}},
 \textbf{Chengcheng Han\textsuperscript{2}},
\\
 \textbf{Qi Gu\textsuperscript{2,$\dagger$}},
 \textbf{Xunliang Cai\textsuperscript{2}},
 \textbf{Xiang Li \textsuperscript{1,$\dagger$}},
 \textbf{Weining Qian \textsuperscript{1}}
\\
\\
 \textsuperscript{1}East China Normal University,
 \textsuperscript{2}Meituan Longcat Team
\\
 \small{
   jiapengzhu@stu.ecnu.edu.cn, xiangli@dase.ecnu.edu.cn, guqi03@meituan.com
 }
}

\begin{document}
\maketitle

\begingroup
\renewcommand{\thefootnote}{} 
\footnotetext{$\ast$ Work done during internship at Meituan.}
\footnotetext{$\dagger$ Corresponding authors.}
\endgroup

\begin{abstract}
Equipping large language models with explicit skills has emerged as a promising paradigm for enabling autonomous agents to solve complex tasks. Agent skills can be inherently divided into general skills for broad cognitive transfer and task-specific skills for dynamic execution. However, existing skill-based reinforcement learning (RL) methods typically force a rigid choice between full externalization, which incurs prohibitive context overhead, and full internalization, which risks overfitting and knowledge conflicts. To address this dilemma, we propose \ours\, a novel agentic RL framework that explicitly differentiates skill treatments by combining general skill internalization with task-specific skill utilization. Driven by a dynamic, difficulty-aware router, \ours\ streams tasks into distinct mastery tiers to apply tailored optimization strategies: it internalizes general skills via privileged distillation to build a cognitive foundation for hard tasks, while using diagnostic probing on easy tasks to penalize shortcuts and enforce specific skill utilization. Experiments on ALFWorld and WebShop demonstrate that \ours\ outperforms both memory-based and skill-based RL baselines, yielding  performance improvements across both in-distribution and out-of-distribution scenarios. The code is available at: \url{https://github.com/JasonZhujp/Skill0_5}.
\end{abstract}


\section{Introduction}

As Large Language Models (LLMs)~\cite{glm,gpt5,longcat} evolve into autonomous problem solvers, they are increasingly entrusted with challenging agentic tasks~\cite{swe,deepresearch,wildclawbench}. To enable agents to master the complex operational logic of real-world tasks, \textit{agent skills} have emerged as a promising solution to break through performance bottlenecks~\cite{survey_forward,agentskill}. A skill encapsulates procedural knowledge into modular, reusable textual directives that codify standard operating procedures and heuristics~\cite{survey_ecosystem}. In practice, these skills are dynamically retrieved and injected into the agent's prompt to explicitly guide it through intricate workflows~\cite{survey_comprehensive}.

While introducing skills via zero-shot prompting offers immediate utility, to further empower agents to robustly navigate complex environments~\cite{metaclaw}, recent research has expanded into skill-based training methods. These methods primarily diverge into two extreme paradigms. One paradigm advocates for \textit{full externalization}, where all skills are maintained as external contextual guidance throughout training and inference~\cite{skillrl, skill1}. Conversely, another line of research explores \textit{full internalization}, aiming to completely assimilate the skills into model parameters~\cite{skill0,skillsd}. In authentic deployment scenarios, however, skill libraries expand dynamically through user contributions, frequently confronting agents with unfamiliar tasks alongside unseen Out-of-Distribution (OOD) task-specific skills~\cite{skillclaw}.

Consequently, both paradigms exhibit notable limitations: \textit{full externalization} imposes severe challenges on LLMs' In-Context Learning (ICL) capabilities~\cite{lost,extern}; as the prompt expands with numerous skills, the excessive length can degrade reasoning and instruction-following performance, especially in long-horizon tasks~\cite{ctx2skill,ruler}. On the other hand, \textit{full internalization} is fundamentally constrained by model capacity~\cite{knowledge_capacity} and potentially introduces knowledge conflict risks~\cite{knowledge_conflict,knowledge_conflict_rag}. Agents may fail to absorb and utilize new instructions when these unfamiliar external skills contradict their internalized skill patterns, leading to execution hallucinations~\cite{work_wild,skillevolver}. Therefore, the efficacy of existing skill-based training approaches in dynamic, real-world environments remains underexplored.

Fundamentally, agentic skills fall into two complementary categories: general and task-specific~\cite{skillrl,skillsbench,skillgraph}. General skills (e.g., meta-reasoning and error recovery) are domain-agnostic but contextually lengthy~\cite{extern}. Conversely, task-specific skills encode granular execution rules that are dynamically updated and susceptible to retrieval noise~\cite{skill0}. However, existing methods treat these categories uniformly, creating a critical dilemma: \textit{fully externalizing} lengthy general heuristics incurs prohibitive context overhead~\cite{agent_memory}, while \textit{fully internalizing} volatile specific skills risks severe overfitting and knowledge conflicts~\cite{avoid_overfitting, evoskill}.
To resolve this, we advocate differentiated treatments: \textbf{internalizing general skills to establish a context-efficient cognitive foundation, while dynamically utilizing plug-and-play task-specific skills to enhance adaptability}, especially in skill-augmented OOD scenarios.

Intuitively, an agent must grasp foundational strategies before exploiting fine-grained rules. Motivated by this, we propose \textbf{\ours}, a unified agentic RL framework that jointly optimizes decoupled general and specific skills based on real-time task mastery. Specifically, a \textit{difficulty-aware router} streams tasks into three tiers for tailored optimization: hard tasks internalize general skills via privileged distillation; medium tasks undergo standard RL to maximize success; and easy tasks employ diagnostic probing to enforce faithful specific skill utilization. Evaluations show \ours\ outperforms the strongest skill-augmented baseline by \textbf{+2.2\%} (ID) and \textbf{+8.5\%} (OOD) across ALFWorld and WebShop.
Our contributions are three-fold:
\begin{itemize}[leftmargin=*,
nosep
]
\item We identify the necessity for differentiated skill treatment in agentic RL, advocating that general skills should be internalized while task-specific skills are dynamically utilized, especially for authentic OOD deployment scenarios.
\item We propose \ours, a novel RL framework featuring an adaptive difficulty-aware router that applies tailored optimization objectives across distinct mastery tiers to jointly internalize and utilize skills.
\item We conduct extensive evaluations on ALFWorld and WebShop under both ID and OOD settings, experimentally demonstrating the effectiveness of joint optimization based on functional skill decoupling.
\end{itemize}

\begin{figure*}[!t]
  \centering
  \includegraphics[width=0.85\textwidth]{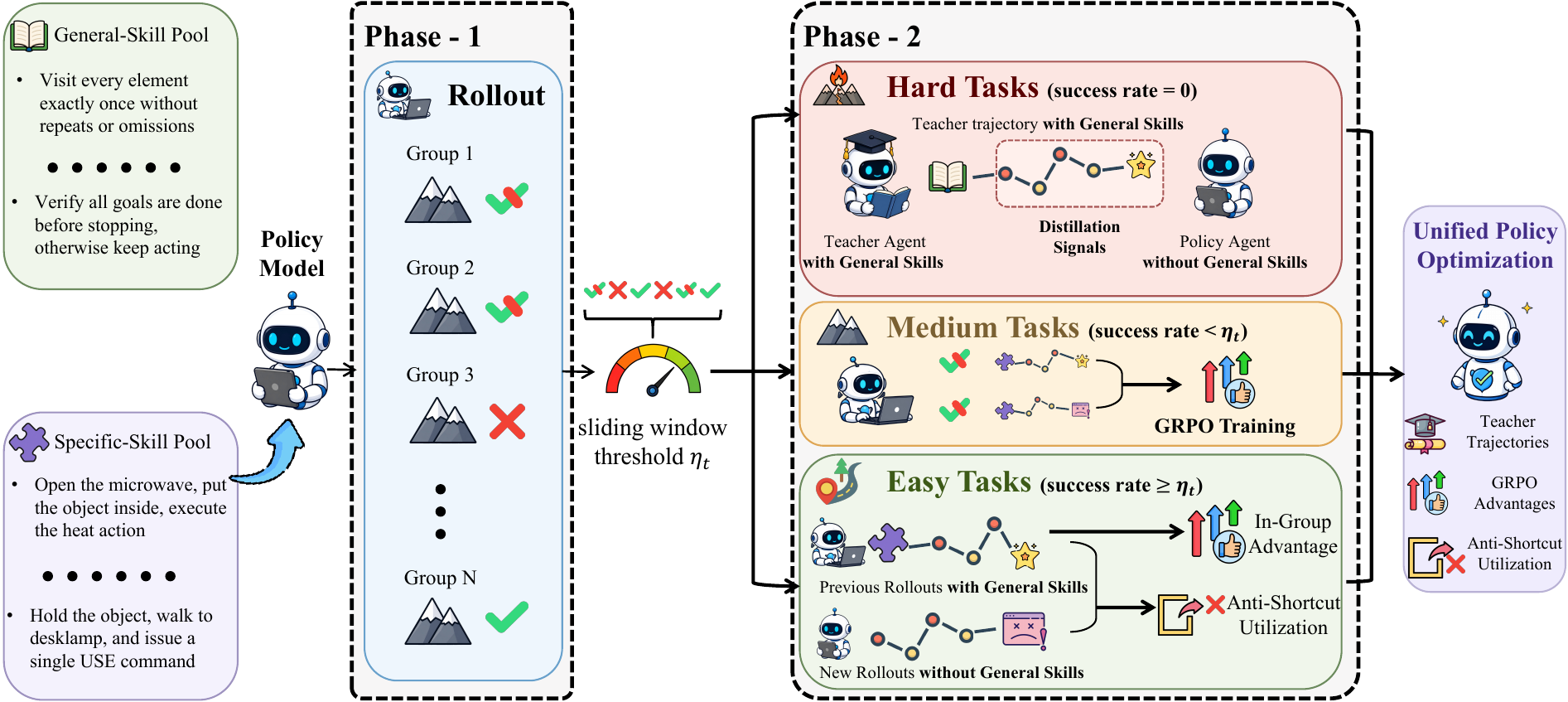}
  \caption{Overall workflow of the \ours\ framework. Skills are explicitly decoupled into general and specific pools. The difficulty-aware router dynamically streams tasks into three tiers for tailored optimization: hard tasks distill general skills, medium tasks apply standard GRPO to improve success rates, and easy tasks employ anti-shortcut probing to provide a utilization advantage.}
  \label{fig:main}
\end{figure*}

\section{Preliminary}

\subsection{Task Formulation}
We consider an agent interacting with a text-based environment $\mathcal{E}$ modeled as a Partially Observable Markov Decision Process (POMDP), designated by the tuple $\mathcal{M} = (\mathcal{S}_{\text{env}}, \mathcal{A}, \mathcal{O}, \mathcal{T}, \Omega, \mathcal{R}_{\text{step}}, \gamma)$. At each observation turn $t$, the agent receives a partial textual observation $o_t \in \mathcal{O}$ that exposes a localized view of the environment state $s_t \in \mathcal{S}_{\text{env}}$. The agent then selects a free-form natural language action $a_t \in \mathcal{A}$, which triggers an environment state transition via $\mathcal{T}(s_{t+1} \mid s_t, a_t)$ and emits the next observation via $\Omega(o_{t+1} \mid s_{t+1}, a_t)$. A complete interactive sequence is captured by an episodic trajectory $\tau = (o_1, a_1, o_2, a_2, \ldots, o_T)$.

Each task is specified by a textual instruction $x$ sampled from a task dataset $\mathcal{D}$. The LLM-based agent, parameterized by $\theta$, generates each action $a_t \sim \pi_\theta(\cdot \mid o_{\le t}, x, c_t)$ conditioned on the interaction history, the task instruction, and an additional context $c_t$ (e.g., skills) injected into the prompt at turn $t$. For simplicity, we abbreviate this execution history as $h_t = (o_{\le t}, x_i)$. Our goal is to optimize the policy parameters $\theta$ to maximize the expected cumulative return across the task distribution:
\[
\max_\theta \mathbb{E}_{x \sim \mathcal{D}, \tau \sim \pi_\theta} \left[ \sum_{t=1}^{T} \gamma^{t-1} \mathcal{R}_{\text{step}}(s_t, a_t) \right]
\]
For outcome-based agentic tasks, this formulation simplifies to a sparse, binary terminal reward $R(\tau) \in \{0, 1\}$, reducing the optimization goal to maximizing the expected success rate $\max_\theta \mathbb{E}[R(\tau)]$. To improve the task success rate, procedural skills are incorporated into the prompt as the \textit{runtime context} $c_t$, which we formalize in the next subsection.

\subsection{Skill Bank and Runtime Context}
Following~\citet{skillrl}, we assume a hierarchical skill bank $\mathcal{S}$ comprising general skills $\mathcal{S}_G$ and specific skills $\mathcal{S}_S$. While $\mathcal{S}_G$ provides universally applicable strategic heuristics, $\mathcal{S}_S$ stores fine-grained execution rules explicitly tied to distinct task domains. At each interaction turn $t$, general skills can be fully provided to the agent due to their broad applicability. For specific skills, which are numerous and semantically fine-grained, an embedding model is used to retrieve a subset most relevant to the task. Let $e_x$ and $e_s$ denote the embeddings of the task instruction $x$ and a candidate skill $s$. The selected specific skill subset $\mathcal{K}_t(x)$ is retrieved via Top-$K$ semantic matching, measured by cosine similarity, across the available specific skill pool:
\begin{equation}
\mathcal{K}_t(x) = \text{TopK}_{s \in \mathcal{S}_S} \big( \cos(e_x, e_s), \; K \big)
\end{equation}

Together with the general skills $\mathcal{S}_G$, this retrieved subset $\mathcal{K}_t(x)$ serves as the candidate guidance for constructing the auxiliary context $c_t$. Different skill-augmented approaches diverge in how they formulate this runtime $c_t$ during training and inference phases:
\begin{itemize}[leftmargin=*,
nosep
]
    \item \textbf{Full Externalization} (e.g., SkillRL~\cite{skillrl}): Involves both general and selected specific skills into the context $c_t = \mathcal{S}_G \cup \mathcal{K}_t(x)$ throughout both phases.
    \item \textbf{Full Internalization} (e.g., SKILL0~\cite{skill0}): Progressively assimilates the full context $c_t = \mathcal{S}_G \cup \mathcal{K}_t(x)$ into model parameters during training to achieve context vacancy $c_t = \emptyset$ at deployment.
    \item \textbf{Hybrid Paradigms:} SLIM~\cite{slim} dynamically maintains $c_t \subset \mathcal{S}$ as an updating active subset during training, and utilizes the final active skill set $c_T \subset \mathcal{S}$ at inference. For our~\ours, we tailor $c_t$ for tasks of varying difficulties during training (elaborated in \ref{sec:method} ), while solely relying on specific skills $c_t = \mathcal{K}_t(x)$ during inference.
\end{itemize}

\subsection{ID and OOD Settings}
\label{sec:id}

We simulate an authentic skill deployment scenario. The complete task domain space is partitioned into ID domains $\mathcal{D}_{\text{id}}$ and OOD domains $\mathcal{D}_{\text{ood}}$, partitioning the entire specific skill pool accordingly into $\mathcal{S}_S^{\text{id}}$ and $\mathcal{S}_S^{\text{ood}}$. The ID tasks are further divided into training splits $\mathcal{X}_{\text{train}}^{\text{id}}$ and validation splits $\mathcal{X}_{\text{val}}^{\text{id}}$. Note that all the general skills $\mathcal{S}_G$ remain globally accessible across all phases, due to their cross-domain applicability. 

During training, the agent encounters ID training tasks $x \sim \mathcal{X}_{\text{train}}^{\text{id}}$ alongside their corresponding ID specific skills $\mathcal{S}_S^{\text{id}}$, whereas OOD tasks and the paired specific skills $\mathcal{S}_S^{\text{ood}}$ remain strictly unobserved. During  evaluation, we assess the agent under two settings: ID evaluation samples tasks from $x \sim \mathcal{X}_{\text{val}}^{\text{id}}$ with retrieval $\mathcal{K}_t(x)$ performed exclusively over $\mathcal{S}_S^{\text{id}}$, while OOD evaluation samples from the unseen $x \sim \mathcal{X}^{\text{ood}}$ with retrieval conducted over the previously unobserved $\mathcal{S}_S^{\text{ood}}$.

Different methods reflect their design principles by how they expose accessible skills at inference. Our philosophy is to fully internalize the strategic essence of $\mathcal{S}_G$ during ID training, and to generalize to unseen tasks by exclusively utilizing plug-and-play specific skills $\mathcal{K}_t(x)$ at evaluation.

\section{Method}\label{sec:method}

Achieving joint skill internalization and utilization requires strategic training design. In cognitive science~\cite{cognitive}, expertise acquisition follows a sequential progression: learners must first construct foundational cognitive schemas before efficiently processing domain-specific rules to prevent cognitive overload. Analogously, an agent cannot effectively utilize task-specific guidance until it has internalized the general logical foundation to interact with the environment.

Motivated by this cognitive progression, we propose \textbf{\ours}, an agentic RL framework that dynamically decouples the optimization towards general and specific skills based on the agent's real-time task mastery. To achieve this, our framework operates in a streamlined two-phase sampling and optimization paradigm, as depicted in Figure~\ref{fig:main}. Specifically, \textbf{Phase-1} (\S\ref{sec:phase1}) executes a difficulty-aware router based on empirical pass rates to stream tasks into three mastery tiers. Subsequently, \textbf{Phase-2} (\S\ref{sec:phase2}) applies tier-tailored optimization: \textit{hard} tasks necessitate the internalization of general heuristics; \textit{medium} tasks prioritize maximizing pass rates; and \textit{easy} tasks ensure that specific skills are genuinely utilized. By providing tailored optimization objectives for each tier, \ours\ promotes the joint internalization and utilization of hierarchical skills.

\subsection{Phase-1: Difficulty-Aware Routing}\label{sec:phase1}
We measure task difficulties using the empirical task pass rate. At step $t$, for each task $x_i$ in batch $\mathcal{B}_t$, we sample $G$ independent trajectories $\tau^{(1)} \sim \pi_\theta(\cdot \mid h_t, c_t^{\text{std}})$ on the \textsc{Standard Prompt}, where $c_t^{\text{std}} = \mathcal{K}_t(x_i)$ ensures only retrieved specific skills are used. This rollout configuration shares the exact same prompt construction as the inference phase. The difficulty of $x_i$ is then evaluated by $p_i = \frac{1}{G} \sum_{g=1}^G R(\tau_i^{(1, g)})$, where $R \in \{0, 1\}$ is the binary environmental outcome. We strictly adhere to the ID training setting formulated in \S\ref{sec:id}, thus omitting the ``id'' superscripts for brevity.

Crucially, these Phase-1 trajectories serve a dual purpose: they act as probing signals to dynamically route the tasks, and are opportunistically reused to support tier-tailored optimization in Phase-2.

Based on the evaluated pass rates, tasks with complete failure, i.e., $p_i = 0$, are directly routed to the \textit{Hard} tier. To further delineate \textit{Medium} from \textit{Easy} tasks, we use a cross-step sliding window average as a dynamic threshold $\eta_t$. This running average is more robust against the limited task amount within a single batch. Given window size $W$ and the batch-level mean $\bar{p}_t = \frac{1}{|\mathcal{B}_t|} \sum_{i \in \mathcal{B}_t} p_i$, the threshold $\eta_t$ averages these means over the past window $[t - \min(W, t) + 1, t]$:
\begin{equation}\label{eq:eta}
\eta_t = \frac{1}{\min(W, t)} \sum_{j=t-\min(W, t)+1}^{t} \bar{p}_j
\end{equation}
Task $x_i$ is treated as \textit{Easy} if $p_i > \eta_t$, and \textit{Medium} otherwise. We formalize this difficulty-aware router $\mathcal{M}(x_i)$ as:
\begin{equation}
\mathcal{M}(x_i) = \begin{cases} 
\text{Hard}, & \text{if } p_i = 0 \\ 
\text{Medium}, & \text{if } 0 < p_i \le \eta_t \\ 
\text{Easy}, & \text{if } p_i > \eta_t 
\end{cases}
\end{equation}

\subsection{Phase-2: Tier-Tailored Optimization}\label{sec:phase2}
Driven by the real-time mastery reflected from Phase-1, the agent now applies targeted optimization objectives for tasks at each tier.

\subsubsection{Hard Tasks: Internalization via Privileged Distillation}\label{sec:hard}
When encountering hard tasks, the agent exposes a lack of foundational reasoning logic. To teach the agent how to think, we introduce the \textsc{Privileged Prompt}, expanding the runtime context to include general heuristics: $c_t^{\text{priv}} = \mathcal{S}_G \cup \mathcal{K}_t(x_i)$. Specifically, we leverage $\mathcal{S}_G$ as privileged information to elicit correct reasoning traces. The agent re-attempts each hard task for $G$ times under this enriched context, performing Phase-2 rollouts as a \textit{teacher}: $\tau^{(2)} \sim \pi_\theta(\cdot \mid h_t, c_t^{\text{priv}})$. These rollouts are filtered for successful trajectories to construct a golden set $\mathcal{T} = \{\tau^{(2)} \mid R(\tau^{(2)}) = 1\}$. Discarding the zero-reward Phase-1 trajectories, we employ teacher forcing to distill this oracle behavior into the student. Specifically, by computing the student's probability distribution along the teacher's successful rollouts $\tau^{(2)} \in \mathcal{T}$, we force the student policy (given only $c_t^{\text{std}}$) to mimic the exact reasoning steps of the teacher (guided by $c_t^{\text{priv}}$). This alignment is optimized via token-level Jensen-Shannon Divergence (JSD), inspired 
by~\citet{hdpo}:
\begin{equation}
\mathcal{L}_{\text{hard}} = \frac{1}{|\mathcal{T}|} \sum_{\tau \in \mathcal{T}} \frac{1}{|\tau|} \sum_{k=1}^{|\tau|} \text{JSD}\big(\text{sg}[\pi_\theta^\text{t}(k)] \;\big\|\; \pi_\theta^\text{s}(k)\big)
\end{equation}
where $\pi_\theta^\text{t}(k) := \pi_\theta(\cdot \mid h_k, c_t^{\text{priv}})$ and $\pi_\theta^\text{s}(k) := \pi_\theta(\cdot \mid h_k, c_t^{\text{std}})$. Here, $\text{sg}[\cdot]$ 
represents the stop-gradient operator, guaranteeing that the student policy actively aligns with the teacher. This enables the agent to handle basic heuristics as if it were guided by $\mathcal{S}_G$ without explicitly conditioning on it, presenting a natural internalization process compatible for inference.

\subsubsection{Medium Tasks: Capability Reinforcement}\label{sec:medium}
For medium tasks whose pass rates fall below the router threshold $\eta_t$, the agent has bypassed the complete cold-start stage but still exhibits substantial room for capability optimization. We directly reuse the Phase-1 trajectories collected during the routing phase, comprising $G$ rollouts for each task. Standard GRPO~\cite{grpo} is applied to maximize the agent's success rate. 

Let the policy ratio for a trajectory $\tau$ at step $k$ be $\rho_k^{(g)} = \frac{\pi_\theta(a_k \mid h_k, c_t^{\text{std}})}{\pi_{\theta_{\text{old}}}(a_k \mid h_k, c_t^{\text{std}})}$. The RL objective $\mathcal{L}_{\text{med}}$ for these medium tasks is formulated as:
\begin{flalign}\label{eq:grpo}
& \mathcal{L}_{\text{medium}} = \frac{1}{G} \sum_{g=1}^G \sum_{k=1}^{|\tau^{(g)}|} \min \Big( \rho_k^{(g)} A_i^{(g)}, & \notag \\
& \hspace{6em} \text{clip}\big(\rho_k^{(g)}, 1-\epsilon, 1+\epsilon\big) A_i^{(g)} \Big) &
\end{flalign}
where $\epsilon$ is the clipping hyperparameter. The advantage $A_i^{(g)}$ is computed via intra-group normalization: $A_i^{(g)} = \frac{R_i^{(g)} - \text{mean}(\mathbf{R}_i)}{\text{std}(\mathbf{R}_i)}$, where $\mathbf{R}_i = \{R_i^{(1)}, \dots, R_i^{(G)}\}$ denotes the rewards for the $G$ trajectories sampled for task $x_i$.

This medium tier functions as the optimization sweet spot~\cite{unveiling}. Through trial and error driven by reward signals, $\mathcal{L}_{\text{med}}$ reinforces the agent's active utilization of specific skills, elevating the sampling efficiency of correct reasoning paths and ultimately maximizing the task success rates.

\subsubsection{Easy Tasks: Anti-Shortcut Utilization}\label{sec:easy}
With the success rate continuously escalating in the easy tier, the policy risks falling into shortcut learning \citep{shortcut}. Rather than genuinely utilizing the retrieved specific skills $\mathcal{K}_t(x_i)$, the agent tends to memorize spurious mappings from task instructions directly to actions. This superficial overfitting severely hurts genuine skill utilization and degrades OOD generalization, where dynamically adapting to unseen specific skills is mandatory.

To penalize such shortcut behaviors, we introduce a counterfactual diagnostic probe: the \textsc{No-Skill Prompt}, where specific skills are deliberately ablated, i.e., $c_t^{none} = \emptyset$. For each easy task $x_i$, we force the agent to perform Phase-2 rollouts under $c_t^{\text{none}}$ to sample $G$ trajectories, and measure the intra-group empirical pass rate $p_i^{\text{none}}$. Crucially, these diagnostic trajectories serve strictly as a counterfactual anchor to isolate the utilization gain, without participating in the policy gradient computation.

We quantify the agent's reliance on specific skills via the utilization gain $u_i = p_i - p_i^{\text{none}}$ where $p_i$ is the original Phase-1 pass rate of the same task $x_i$ conditioned on $c_t^{std}$. Intuitively, this gain captures the causal impact of the specific skills on task success. A robust agent equipped with necessary skills should strictly outperform its unguided counterpart. When $u_i$ shrinks or becomes negative, it exposes the agent's behavior of bypassing the external guidance.

To optimize for this reliance, we apply a sliding window to track the mean utilization gain over the recent $W$ steps, denoted as $\bar{u}_t$. By treating $\bar{u}_t$ as a dynamic anchor, we naturally construct an auxiliary task-level utilization advantage $A_i^u$ for the tasks in the current batch: 
\begin{equation}
    A_i^u = \frac{u_i - \overline{u}_t}{\sigma_u}
\end{equation}
where $\sigma_u$ is the batch-level standard deviation of $(u_i - \overline{u}_t)$. Unlike the standard intra-group advantage $A_i^{(g)}$ which performs zero-mean normalization to evaluate the \textit{relative} quality among trajectories, $A_i^u$ serves as a \textit{global task-level modulator}. It shifts the entire advantage landscape of the task. The composite advantage for the $g$-th rollout (sampled from Phase-1 under $c_t^{std}$) is thus formulated as:
\begin{equation}
    \hat{A}_i^{(g)} = \underbrace{A_i^{(g)}}_{\text{Trajectory-level quality}} + \underbrace{A_i^u}_{\text{Task-level utilization}}
\end{equation}
If a task exposes shortcut learning ($u_i < \overline{u}_t$), the negative offset $A_i^u$ globally suppresses the optimization gradients for this task, penalizing the distribution of actions that bypass specific skills.
Finally, the objective $\mathcal{L}_{easy}$ is optimized by substituting the standard $A_i^{(g)}$ with the composite advantage $\hat{A}_i^{(g)}$ into the identical GRPO framework.

\begin{table*}[t]
\centering
\footnotesize
\setlength{\tabcolsep}{4.0pt}
\renewcommand{\arraystretch}{0.9}

\resizebox{0.85\textwidth}{!}{%
\begin{tabular}{lccccc|ccccc}
\toprule

\multirow{2}{*}{\textbf{Method}} 
& \multicolumn{5}{c|}{\textbf{ID}} 
& \multicolumn{5}{c}{\textbf{OOD}} \\

\cmidrule(lr){2-6}
\cmidrule(lr){7-11}

& \textbf{Pick} 
& \textbf{Cool} 
& \textbf{Clean} 
& \textbf{Avg.} 
& \textcolor{gray}{\textbf{Rank $\downarrow$}}
& \textbf{Look} 
& \textbf{Heat} 
& \textbf{Pick2} 
& \textbf{Avg.} 
& \textcolor{gray}{\textbf{Rank $\downarrow$}} \\

\midrule

\multicolumn{11}{l}{\emph{Prompt-based Methods}} \\

Zero-shot 
& 28.6 & 12.0 & 18.5 & 20.7 & \textcolor{gray}{17.2}
& 38.5 & 12.5 & 12.5 & 18.9 & \textcolor{gray}{15.5} \\

Few-shot 
& 62.9 & 44.0 & 63.0 & 57.5 & \textcolor{gray}{12.0}
& 46.2 & 31.2 & 8.3 & 24.5 & \textcolor{gray}{12.8} \\

ReAct 
& 71.4 & 28.0 & 33.3 & 47.1 & \textcolor{gray}{13.3}
& 46.2 & 18.8 & 12.5 & 22.6 & \textcolor{gray}{12.8} \\

Reflexion 
& 85.7 & 44.0 & 44.4 & 60.9 & \textcolor{gray}{10.3}
& 46.2 & 31.3 & 29.2 & 34.0 & \textcolor{gray}{8.7} \\

Mem0 
& 54.3 & 4.0 & 18.5 & 28.7 & \textcolor{gray}{17.5}
& 38.5 & 18.8 & 4.2 & 17.0 & \textcolor{gray}{15.7} \\

ExpeL 
& 80.0 & 44.0 & 66.7 & 65.5 & \textcolor{gray}{9.5}
& 46.2 & 18.8 & 20.8 & 18.9 & \textcolor{gray}{11.2} \\

MemP 
& 65.7 & 12.0 & 33.3 & 40.2 & \textcolor{gray}{15.3}
& 46.2 & 37.5 & 12.5 & 28.3 & \textcolor{gray}{11.0} \\

SimpleMem 
& 71.4 & 16.0 & 44.4 & 47.1 & \textcolor{gray}{13.3}
& 53.8 & 18.8 & 20.8 & 28.3 & \textcolor{gray}{9.5} \\

\midrule

\multicolumn{11}{l}{\emph{RL-based Methods}} \\

RLOO 
& \cellcolor{second}\underline{91.4}
& 80.0
& 81.5
& 85.1
& \textcolor{gray}{4.3}

& 61.5
& 56.3
& 20.8
& 41.5
& \textcolor{gray}{5.8} \\

GRPO 
& 80.0

& 72.0

& \cellcolor{second}\underline{88.9}

& 80.5

& \textcolor{gray}{6.2}

& \cellcolor{best}\textbf{76.9}

& 56.3

& 16.7

& 43.4

& \textcolor{gray}{5.5} \\

\midrule

\multicolumn{11}{l}{\emph{Memory-Augmented RL Methods}} \\

MemRL 
& 74.3 & 12.0 & 55.6 & 50.6 & \textcolor{gray}{12.7}
& 46.2 & 12.5 & \cellcolor{second}\underline{45.8} & 35.8 & \textcolor{gray}{9.7} \\

EvolveR 
& 88.6 & 52.0 & 81.5 & 75.9 & \textcolor{gray}{6.2}
& 46.2 & 6.2 & \cellcolor{best}\textbf{50.0} & 35.8 & \textcolor{gray}{10.0} \\

Mem0+GRPO 
& 65.7 & 20.0 & 51.9 & 48.3 & \textcolor{gray}{13.2}
& 23.1 & 6.2 & 20.8 & 17.0 & \textcolor{gray}{15.0} \\

SimpleMem+GRPO 
& 85.7 & 52.0 & 70.3 & 71.3 & \textcolor{gray}{7.7}
& 61.5 & 43.8 & 41.7 & \cellcolor{second}\underline{47.2} & \cellcolor{second}\textcolor{gray}{\underline{4.5}} \\

\midrule

\multicolumn{11}{l}{\emph{Skill-Augmented RL Methods}} \\

SkillRL 
& \cellcolor{second}\underline{91.4}
& \cellcolor{second}\underline{84.0}
& \cellcolor{best}\textbf{96.3}
& \cellcolor{second}\underline{90.8}
& \cellcolor{second}\textcolor{gray}{\underline{2.7}}

& \cellcolor{second}\underline{69.2}
& \cellcolor{second}\underline{75.0}
& 12.5
& 45.3
& \textcolor{gray}{6.3} \\

SKILL0 
& \cellcolor{best}\textbf{94.3}
& 76.0
& 81.5
& 85.1
& \textcolor{gray}{3.8}

& 46.2
& 50.0
& 29.2
& 39.6
& \textcolor{gray}{7.5} \\

SLIM 
& \cellcolor{second}\underline{91.4}
& \cellcolor{second}\underline{84.0}
& 70.4
& 82.8
& \textcolor{gray}{4.5}

& 53.8
& 31.3
& 29.2
& 35.8
& \textcolor{gray}{7.0} \\

\textbf{\ours{}}

& \cellcolor{best}\textbf{94.3}

& \cellcolor{best}\textbf{88.0}

& \cellcolor{best}\textbf{96.3}

& \cellcolor{best}\textbf{93.1}

& \cellcolor{best}\textcolor{gray}{\textbf{1.3}}

& \cellcolor{second}\underline{69.2}

& \cellcolor{best}\textbf{87.5}

& 33.3

& \cellcolor{best}\textbf{58.5}

& \cellcolor{best}\textcolor{gray}{\textbf{2.5}} \\

\bottomrule
\end{tabular}%
}

\caption{
Performance comparison on ALFWorld under ID and OOD task settings.
\colorbox[rgb]{0.79,0.86,0.97}{\textbf{Best}} and
\colorbox[rgb]{0.91,0.94,0.98}{\underline{second-best}}
results in each column are highlighted, respectively.
Average \textcolor{gray}{Rank} is computed across all evaluated settings.
}

\label{tab:alfworld_results}

\end{table*}

\paragraph{Overall Objective.} Ultimately, the global optimization objective of \ours\ is formulated as the joint aggregation of the tier-specific losses:
\begin{equation}
    \mathcal{L} = \mathcal{L}_{hard} + \mathcal{L}_{medium} + \mathcal{L}_{easy}
\end{equation}
For any single task $x_i$ within a training batch, these optimization signals are mutually exclusive due to the routing boundaries. This dynamic routing mechanism establishes a structured curriculum synchronized with the agent's real-time mastery dynamics. By allocating tailored learning objectives based on real-time task mastery, \ours\ achieves joint optimization of foundational reasoning internalization and task-specific guidance utilization within a unified RL framework. The full procedure is summarized in Algorithm~\ref{alg:skill05}.

\section{Experiments}

\subsection{Experimental Setup}

\paragraph{Environments and ID/OOD Partition.} 
We evaluate our framework on two multi-turn interactive benchmarks that offer clear domain segmentation, enabling a rigorous study of OOD generalization.
\begin{itemize}[leftmargin=*, 
nosep
]
    \item 
    \textbf{ALFWorld}~\cite{alfworld} is a text-based embodied environment where agents complete household tasks through natural language actions. We evaluate on its six canonical task types.
    We designate \{Pick, Cool, Clean\} as ID and \{Look, Heat, Pick2\} as OOD domains. 
    \item 
    \textbf{WebShop}~\cite{webshop} is a web-based shopping environment where agents search for products and make purchases matching user instructions. 
    We split product categories into ID = \{Apparel, Electronics, Footwear, Other\} and OOD = \{Accessories, Beauty \& Health, Home Decor\} domains following a balanced protocol detailed in Appendix~\ref{sec:split}.
    The OOD categories exhibit distinct attribute vocabularies and product matching heuristics entirely absent from training.
\end{itemize}

For \textbf{agent skills}, we adopt the hierarchical Skill Bank proposed by \citet{skillrl} as our foundational skill set. The library comprises 12 and 15 general skills for ALFWorld and WebShop, respectively, while each task domain maintains around 5 task-specific skills.

\paragraph{Baselines.} 
We compare \ours\ against diverse spectrum of methods: (1) \textbf{Prompt-based Methods}: Zero-shot and Few-shot prompting. (2) \textbf{Prompt-based Agentic or Memory-based Methods}: ReAct~\cite{react} and Reflexion~\cite{reflexion}, which rely on in-context prompting for multi-step reasoning, alongside Mem0~\cite{mem0}, ExpeL~\cite{expel}, MemP~\cite{memp}, and SimpleMem~\cite{simplemem}, which utilize external experience pools to guide behavior without parameter updates. (3) \textbf{RL-based Methods}: Group-based RL algorithms such as RLOO~\cite{rloo} and GRPO~\cite{grpo}. (4) \textbf{Memory-Augmented RL}: MemRL~\cite{memrl}, EvolveR~\cite{evolver}, Mem0+GRPO, and SimpleMem+GRPO, which integrate persistent memory directly into RL optimization. (5) \textbf{Skill-Augmented RL}: SkillRL~\cite{skillrl}, SKILL0~\cite{skill0}, and SLIM~\cite{slim}, which represent the current frontier of skill-based agent training.

The implementation details and hyperparameter configurations are provided in Appendix~\ref{sec:immplement} and~\ref{sec:hyper}. 

\subsection{Main Results}

Table~\ref{tab:alfworld_results} and Table~\ref{tab:webshop_results} report the comprehensive success rates across distinct domains of ALFWorld and WebShop, detailing both the task-specific performance and the aggregated ID and OOD averages. We draw the following key observations:

\textbf{\ours\ achieves the highest overall performance.} Across all settings, \ours\ establishes a decisive performance advantage over the full spectrum of baselines. Compared to the strongest skill-augmented baseline, SkillRL, \ours\ achieves absolute improvements of \textbf{+2.3\%} (ID) and \textbf{+13.2\%} (OOD) on ALFWorld, with consistent gains of \textbf{+2.1\%} and \textbf{+3.9\%} on the respective splits of WebShop. These results confirm that our joint internalization and utilization framework ensures steady ID progress while unlocking significant generalization leaps in OOD scenarios.

\textbf{Prompt-based methods establish a performance floor.} Although prompting with skills or memory (e.g., Few-shot, ReAct) improves upon zero-shot baselines, these methods significantly lag behind \ours, trailing by an average of over 45\% on ALFWorld and 28\% on WebShop. This profound gap highlights that relying solely on in-context learning is insufficient to effectively synergize with external skills~\cite{work_wild}.

\textbf{Memory-based methods are bottlenecked by context noise.} Methods integrating trajectory storage into the prompt are highly sensitive to memory quality. For instance, SimpleMem+GRPO achieves 47.2\% on ALFWorld OOD thanks to its sophisticated memory management, whereas Mem0+GRPO collapses to 17.0\%. However, even the strongest memory-augmented methods fail to surpass skill-augmented approaches. This discrepancy arises because raw memory retrieval tends to inject overly detailed context noise, whereas effective knowledge transfer demands high-level procedural abstractions and reusable heuristics~\cite{skillrl}.

\begin{figure*}[!t]
  \centering
  \includegraphics[width=\textwidth]{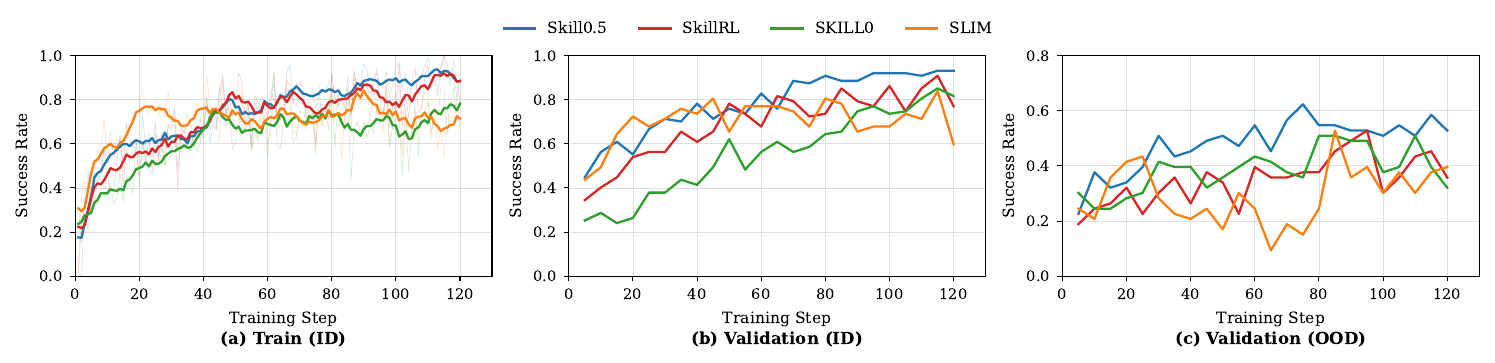}
  \caption{Success rates across the training and validation sets on ALFWorld, compared to skill-based RL baselines.}
  \label{fig:alfworld_dynamics}
\end{figure*}

\begin{figure}[!t]
    \centering
    \includegraphics[width=\linewidth]{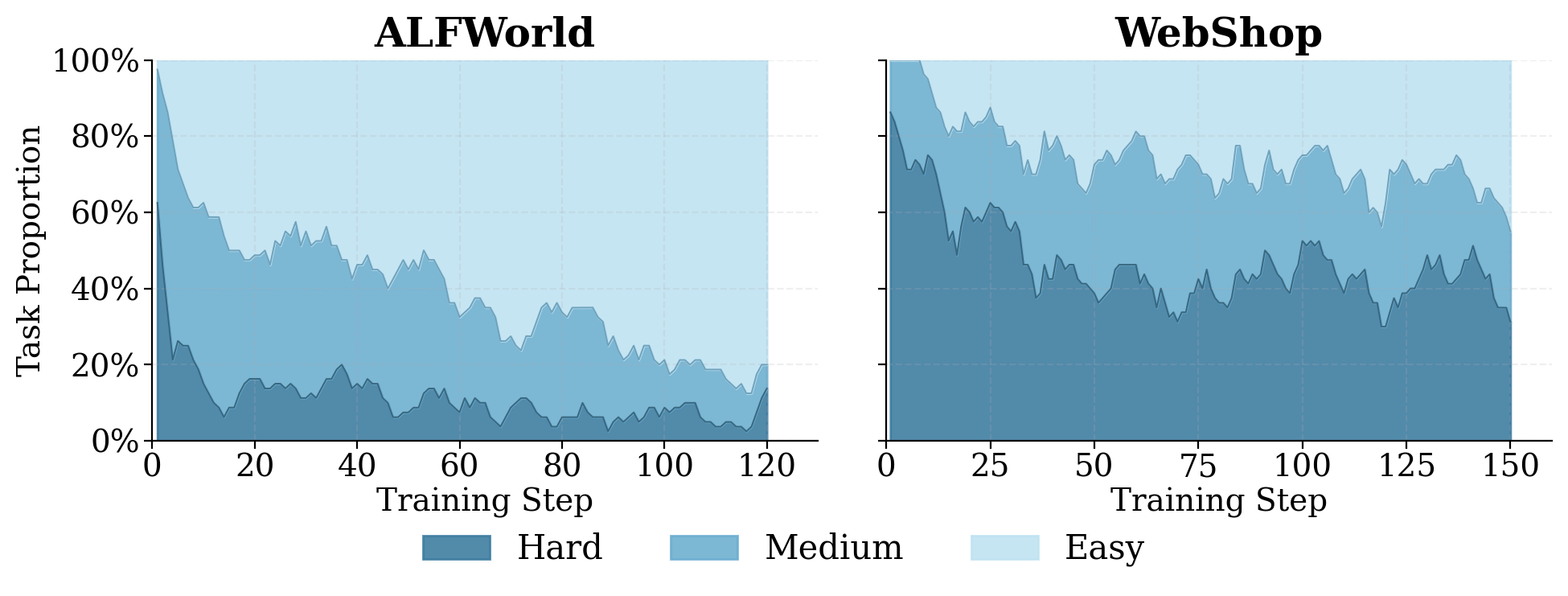}
    \caption{Dynamic distribution of task difficulties allocated by our difficulty-aware router.}
    \label{fig:router}
    \vspace{-2mm}
\end{figure}

\subsection{Training Dynamics}
\label{sec:dynamics}

Here, we investigate the training dynamics to elucidate why our approach consistently surpasses existing skill-augmented RL baselines. Figure~\ref{fig:alfworld_dynamics} illustrates the success rate curves on the training and validation (ID and OOD) sets of ALFWorld across training steps. Concurrently, Figure~\ref{fig:router} tracks the dynamic proportions of hard, medium, and easy tasks allocated by our difficulty-aware router.

\textbf{Early-stage training: Overcoming the zero-gradient dilemma for rapid internalization.}
Initially, hard tasks dominate the distribution, causing zero reward variance and vanishing advantages that entirely eliminate gradient signals~\cite{unveiling}. Our difficulty-aware router resolves this by triggering privileged distillation as a surrogate gradient source, which breaks the exploration deadlock and swiftly steers optimization, enabling \ours\ to achieve markedly faster initial ascent compared to baselines.

\textbf{Mid-to-late training: Anti-shortcut utilization driving robust OOD generalization.}
As training progresses, easy tasks gradually dominate, creating a phase prone to shortcut learning. Owing to our anti-shortcut diagnostic probing, \ours\ maintains steady growth on training and ID validation sets while achieving consistent, un-decaying improvements on OOD tasks. This demonstrates that our method genuinely learns to utilize novel skills, rather than overfitting to bypass skill guidance.

\textbf{Limitations of baseline methods.}
SkillRL exhibits overfitting: despite a surging training success rate, its performance on validation and OOD tasks declines in later stages. This indicates a collapse into shortcut learning, where the agent memorizes domain-specific actions at the expense of adaptability. Conversely, SKILL0 relies on pure internalization but is consistently suppressed on OOD tasks, confirming that a fully internalized policy is too rigid to integrate novel task-skill pairs. Finally, SLIM suffers from severe oscillations due to its dynamic retirement mechanism prematurely discarding general skills. Losing this foundational reasoning halts progress on hard tasks, and the resulting disjointed active skill set inevitably mismatches with OOD tasks.

\subsection{Ablation Study} 
To isolate the contributions of our tier-tailored optimization, we evaluate two ablated variants on ALFWorld: \textbf{Internalize-Only} (retains distillation for hard tasks; standard GRPO elsewhere) and \textbf{Utilize-Only} (retains contrastive utilization for easy tasks; standard GRPO elsewhere).

\begin{table}[h!]
\centering
\setlength{\tabcolsep}{3pt}
\resizebox{0.70\linewidth}{!}{%
\begin{tabular}{lcc}
\toprule
\textbf{Variant} & \textbf{ID Avg.} & \textbf{OOD Avg.} \\
\midrule
Internalize-Only & 89.6 & 52.8 \\
Utilize-Only     & 85.1 & 50.9 \\
\midrule
\textbf{\ours} & \textbf{93.1} & \textbf{58.5} \\
\bottomrule
\end{tabular}
}
\caption{Ablation study on ALFWorld. We report the average success rates across ID and OOD task splits.}
\label{tab:ablation}
\end{table}

Table~\ref{tab:ablation} details the results, revealing two critical insights:
(1) \textbf{Internalization is a strict prerequisite for overall capability.} The \textit{Utilize-Only} variant suffers a catastrophic decline across both ID and OOD splits. Although the adaptive router still allocates tasks to the easy tier based on relative performance, the contrastive utilization objective becomes futile: lacking basic competence, the agent fails regardless of whether specific skills are provided, yielding a negligible contrastive advantage that severely stalls the entire training process.
(2) \textbf{Utilization unlocks peak OOD adaptability.} The \textit{Internalize-Only} variant experiences a moderate performance drop. While the internalized general skills successfully provide a robust reasoning baseline (preventing a complete collapse), the absence of the contrastive utilization loss prevents the model from faithfully grounding its actions in novel, task-specific OOD guidance, limiting its peak generalization.

Together, these results validate the necessity of our joint design: internalization builds the indispensable reasoning foundation, upon which utilization further maximizes OOD adaptability.

\subsection{Case Study}

We conduct qualitative trajectory analysis on ALFWorld OOD tasks to diagnose why each baseline fails (full details in Appendix~\ref{sec:case_study}). We identify three distinct failure mechanisms: SkillRL suffers from \textit{Contextual Interference}, where OOD-specific skills 
are overridden by ID-trained habitual associations from the dominant general context; SKILL0 exhibits \textit{Parametric Knowledge Conflict}, where internalized ID templates produce actions incompatible with OOD procedures despite the agent referencing the correct skill in its reasoning; and SLIM demonstrates \textit{Behavioral Collapse}, progressively degenerating from success
to task hallucination 
after prematurely retiring general skills. In all cases, \ours\ succeeds by internalizing general reasoning into parameters while faithfully executing novel OOD-specific skills, validating our thesis that the two skill types require differentiated treatments.

\section{Conclusion}
In this paper, we tackled the skill-treatment dilemma for LLM agents, where existing RL methods typically suffered from a rigid choice between full context externalization and indiscriminate internalization.
To resolve this, we introduced \ours, a unified framework that achieves joint optimization by explicitly differentiating the treatment of decoupled general and task-specific skills. Driven by a dynamic, difficulty-aware router, \ours\ orchestrated optimization strategies tailored to different task tiers: it compressed general skills via privileged distillation to build a cognitive foundation for hard tasks, while enforcing specific skill utilization via contrastive advantage probing on easy tasks. Extensive experiments on the ALFWorld and WebShop benchmarks demonstrated that \ours\ significantly outperformed prompt-, memory-, and skill-based baselines, validating that our framework yields moderate improvements on ID tasks while substantially enhancing the agent's generalization capability across practical deployment settings involving unseen OOD tasks.

\clearpage

\section*{Limitations}
While our framework is validated on text-based interactive environments, the principle of differentiated skill treatment is broadly applicable. In future work, we plan to extend \ours\ to more complex domains such as code generation, multi-modal environments, and open-ended web navigation, as well as to settings with longer horizons and larger action spaces.

\section*{Ethical Considerations}

This work focuses on improving the reasoning and generalization capabilities of LLM-based agents within simulated environments (ALFWorld and WebShop). We identify no direct ethical risks arising from our research. All experiments are conducted in controlled, sandboxed settings with no real-world deployment or interaction with human users. The benchmarks used are publicly available and do not involve personally identifiable information or sensitive content. Our method does not introduce new data collection procedures, and all training data is derived from synthetic environment interactions. We acknowledge that advances in autonomous agent capabilities carry broader societal implications; however, the household and e-commerce simulation domains studied here pose minimal risk of misuse.



\bibliography{custom}

@article{swe,
  title={SWE-Cycle: Benchmarking Code Agents across the Complete Issue Resolution Cycle},
  author={Guan, Hao and Fu, Lingyue and Zhang, Shao and Zhu, Yaoming and Zhang, Kangning and Qiu, Lin and Cai, Xunliang and Cao, Xuezhi and Liu, Weiwen and Zhang, Weinan and others},
  journal={arXiv preprint arXiv:2605.13139},
  year={2026}
}

@article{deepresearch,
  title={Deepresearch bench: A comprehensive benchmark for deep research agents},
  author={Du, Mingxuan and Xu, Benfeng and Zhu, Chiwei and Wang, Xiaorui and Mao, Zhendong},
  journal={arXiv preprint arXiv:2506.11763},
  year={2025}
}

@article{wildclawbench,
  title={WildClawBench: A Benchmark for Real-World, Long-Horizon Agent Evaluation},
  author={Ding, Shuangrui and Dai, Xuanlang and Xing, Long and Ding, Shengyuan and Liu, Ziyu and JingYi, Yang and Yang, Penghui and Zhang, Zhixiong and Wei, Xilin and Fang, Xinyu and others},
  journal={arXiv preprint arXiv:2605.10912},
  year={2026}
}

@article{skillsbench,
  title={SkillsBench: Benchmarking how well agent skills work across diverse tasks},
  author={Li, Xiangyi and Chen, Wenbo and Liu, Yimin and Zheng, Shenghan and Chen, Xiaokun and He, Yifeng and Li, Yubo and You, Bingran and Shen, Haotian and Sun, Jiankai and others},
  journal={arXiv preprint arXiv:2602.12670},
  year={2026}
}

@article{agentskill,
  title={Agent skills: A data-driven analysis of claude skills for extending large language model functionality},
  author={Ling, George and Zhong, Shanshan and Huang, Richard},
  journal={arXiv preprint arXiv:2602.08004},
  year={2026}
}

@article{survey_comprehensive,
  title={A Comprehensive Survey on Agent Skills: Taxonomy, Techniques, and Applications},
  author={Zhou, Yingli and Shu, Wang and Su, Yaodong and Du, Wenchuan and Fang, Yixiang and Lin, Xuemin},
  journal={arXiv preprint arXiv:2605.07358},
  year={2026}
}

@article{survey_forward,
  title={Agent skills for large language models: Architecture, acquisition, security, and the path forward},
  author={Xu, Renjun and Yan, Yang},
  journal={arXiv preprint arXiv:2602.12430},
  year={2026}
}

@article{survey_ecosystem,
  title={Organizing, orchestrating, and benchmarking agent skills at ecosystem scale},
  author={Li, Hao and Mu, Chunjiang and Chen, Jianhao and Ren, Siyue and Cui, Zhiyao and Zhang, Yiqun and Bai, Lei and Hu, Shuyue},
  journal={arXiv preprint arXiv:2603.02176},
  year={2026}
}

@article{glm,
  title={Glm-5: from vibe coding to agentic engineering},
  author={Zeng, Aohan and Lv, Xin and Hou, Zhenyu and Du, Zhengxiao and Zheng, Qinkai and Chen, Bin and Yin, Da and Ge, Chendi and Huang, Chenghua and Xie, Chengxing and others},
  journal={arXiv preprint arXiv:2602.15763},
  year={2026}
}

@article{gpt5,
  title={Openai gpt-5 system card},
  author={Singh, Aaditya and Fry, Adam and Perelman, Adam and Tart, Adam and Ganesh, Adi and El-Kishky, Ahmed and McLaughlin, Aidan and Low, Aiden and Ostrow, AJ and Ananthram, Akhila and others},
  journal={arXiv preprint arXiv:2601.03267},
  year={2025}
}

@article{longcat,
  title={Longcat-flash-thinking-2601 technical report},
  author={Team, Meituan LongCat and Gui, Anchun and Li, Bei and Tao, Bingyang and Zhou, Bole and Chen, Borun and Zhang, Chao and Gao, Chen and Zhang, Chen and Han, Chengcheng and others},
  journal={arXiv preprint arXiv:2601.16725},
  year={2026}
}

@article{metaclaw,
  title={MetaClaw: Just Talk--An Agent That Meta-Learns and Evolves in the Wild},
  author={Xia, Peng and Chen, Jianwen and Yang, Xinyu and Tu, Haoqin and Liu, Jiaqi and Xiong, Kaiwen and Han, Siwei and Qiu, Shi and Ji, Haonian and Zhou, Yuyin and others},
  journal={arXiv preprint arXiv:2603.17187},
  year={2026}
}

@article{work_wild,
  title={How well do agentic skills work in the wild: Benchmarking llm skill usage in realistic settings},
  author={Liu, Yujian and Ji, Jiabao and An, Li and Jaakkola, Tommi and Zhang, Yang and Chang, Shiyu},
  journal={arXiv preprint arXiv:2604.04323},
  year={2026}
}

@article{skillgraph,
  title={SkillGraph: Skill-Augmented Reinforcement Learning for Agents via Evolving Skill Graphs},
  author={Li, Xiaoyuan and Li, Moxin and Bao, Keqin and Ma, Yubo and Wang, Wenjie and Liu, Dayiheng and Feng, Fuli},
  journal={arXiv preprint arXiv:2605.12039},
  year={2026}
}

@article{skillrl,
  title={Skillrl: Evolving agents via recursive skill-augmented reinforcement learning, 2026},
  author={Xia, Peng and Chen, Jianwen and Wang, Hanyang and Liu, Jiaqi and Zeng, Kaide and Wang, Yu and Han, Siwei and Zhou, Yiyang and Zhao, Xujiang and Chen, Haifeng and others},
  journal={URL https://arxiv. org/abs/2602.08234},
  year={2026}
}

@article{skill0,
  title={Skill0: In-context agentic reinforcement learning for skill internalization},
  author={Lu, Zhengxi and Yao, Zhiyuan and Wu, Jinyang and Han, Chengcheng and Gu, Qi and Cai, Xunliang and Lu, Weiming and Xiao, Jun and Zhuang, Yueting and Shen, Yongliang},
  journal={arXiv preprint arXiv:2604.02268},
  year={2026}
}

@article{skillsd,
  title={Skill-sd: Skill-conditioned self-distillation for multi-turn llm agents},
  author={Wang, Hao and Wang, Guozhi and Xiao, Han and Zhou, Yufeng and Pan, Yue and Wang, Jichao and Xu, Ke and Wen, Yafei and Ruan, Xiaohu and Chen, Xiaoxin and others},
  journal={arXiv preprint arXiv:2604.10674},
  year={2026}
}

@article{skill1,
  title={Skill1: Unified Evolution of Skill-Augmented Agents via Reinforcement Learning},
  author={Shi, Yaorui and Chen, Yuxin and Lu, Zhengxi and Miao, Yuchun and Liu, Shugui and Gu, Qi and Cai, Xunliang and Wang, Xiang and Zhang, An},
  journal={arXiv preprint arXiv:2605.06130},
  year={2026}
}

@article{slim,
  title={Dynamic Skill Lifecycle Management for Agentic Reinforcement Learning},
  author={Shen, Junhao and Zhang, Teng and Zhao, Xiaoyan and Cheng, Hong},
  journal={arXiv preprint arXiv:2605.10923},
  year={2026}
}

@article{skillclaw,
  title={Skillclaw: Let skills evolve collectively with agentic evolver},
  author={Ma, Ziyu and Yang, Shidong and Ji, Yuxiang and Wang, Xucong and Wang, Yong and Hu, Yiming and Huang, Tongwen and Chu, Xiangxiang},
  journal={arXiv preprint arXiv:2604.08377},
  year={2026}
}

@article{lost,
  title={Lost in the middle: How language models use long contexts},
  author={Liu, Nelson F and Lin, Kevin and Hewitt, John and Paranjape, Ashwin and Bevilacqua, Michele and Petroni, Fabio and Liang, Percy},
  journal={Transactions of the association for computational linguistics},
  volume={12},
  pages={157--173},
  year={2024}
}

@article{ctx2skill,
  title={From Context to Skills: Can Language Models Learn from Context Skillfully?},
  author={Si, Shuzheng and Zhao, Haozhe and Lei, Yu and Wang, Qingyi and Chen, Dingwei and Wang, Zhitong and Wang, Zhenhailong and Luo, Kangyang and Wang, Zheng and Chen, Gang and others},
  journal={arXiv preprint arXiv:2604.27660},
  year={2026}
}

@article{ruler,
  title={RULER: What's the real context size of your long-context language models?},
  author={Hsieh, Cheng-Ping and Sun, Simeng and Kriman, Samuel and Acharya, Shantanu and Rekesh, Dima and Jia, Fei and Zhang, Yang and Ginsburg, Boris},
  journal={arXiv preprint arXiv:2404.06654},
  year={2024}
}

@article{extern,
  title={Externalization in llm agents: A unified review of memory, skills, protocols and harness engineering},
  author={Zhou, Chenyu and Chai, Huacan and Chen, Wenteng and Guo, Zihan and Shan, Rong and Song, Yuanyi and Xu, Tianyi and Yang, Yingxuan and Yu, Aofan and Zhang, Weiming and others},
  journal={arXiv preprint arXiv:2604.08224},
  year={2026}
}

@article{skillevolver,
  title={SkillEvolver: Skill Learning as a Meta-Skill},
  author={Zhang, Genrui and Zhu, Erle and Zhou, Jinfeng and Jia, Caiyan and Wang, Hongning},
  journal={arXiv preprint arXiv:2605.10500},
  year={2026}
}

@inproceedings{knowledge_conflict,
  title={Knowledge conflicts for llms: A survey},
  author={Xu, Rongwu and Qi, Zehan and Guo, Zhijiang and Wang, Cunxiang and Wang, Hongru and Zhang, Yue and Xu, Wei},
  booktitle={Proceedings of the 2024 Conference on Empirical Methods in Natural Language Processing},
  pages={8541--8565},
  year={2024}
}

@inproceedings{knowledge_capacity,
  author = {{Allen-Zhu}, Zeyuan and Li, Yuanzhi},
  title = {{Physics of Language Models: Part 3.3, Knowledge Capacity Scaling Laws}},
  booktitle = {Proceedings of the 13th International Conference on Learning Representations},
  series = {ICLR~'25},
  month = apr,
  year = 2025,
  note = {Full version available at \url{https://ssrn.com/abstract=5250617}}
}

@inproceedings{knowledge_conflict_rag,
  title={Astute rag: Overcoming imperfect retrieval augmentation and knowledge conflicts for large language models},
  author={Wang, Fei and Wan, Xingchen and Sun, Ruoxi and Chen, Jiefeng and Arik, Sercan O},
  booktitle={Proceedings of the 63rd Annual Meeting of the Association for Computational Linguistics (Volume 1: Long Papers)},
  pages={30553--30571},
  year={2025}
}

@article{agent_memory,
  title={Agent Skills from the Perspective of Procedural Memory: A Survey},
  author={Wu, Yaxiong and Zhang, Yongyue},
  journal={Authorea Preprints},
  year={2026},
  publisher={Authorea}
}

@article{avoid_overfitting,
  title={Scaling coding agents via atomic skills},
  author={Ma, Yingwei and Liu, Yue and Yang, Xinlong and Li, Yanhao and Fu, Kelin and Miao, Yibo and Xie, Yuchong and Wang, Zhexu and Cheung, Shing-Chi},
  journal={arXiv preprint arXiv:2604.05013},
  year={2026}
}

@article{evoskill,
  title={Evoskill: Automated skill discovery for multi-agent systems},
  author={Alzubi, Salaheddin and Provenzano, Noah and Bingham, Jaydon and Chen, Weiyuan and Vu, Tu},
  journal={arXiv preprint arXiv:2603.02766},
  year={2026}
}

@article{arise,
  title={Arise: Agent reasoning with intrinsic skill evolution in hierarchical reinforcement learning},
  author={Li, Yu and Miao, Rui and Qi, Zhengling and Lan, Tian},
  journal={arXiv preprint arXiv:2603.16060},
  year={2026}
}

@article{hintsd,
  title={HINT-SD: Targeted Hindsight Self-Distillation for Long-Horizon Agents},
  author={Yeo, Woongyeng and Choi, Yumin and Ki, Taekyung and Hwang, Sung Ju},
  journal={arXiv preprint arXiv:2605.17873},
  year={2026}
}

@article{sdar,
  title={Self-Distilled Agentic Reinforcement Learning},
  author={Lu, Zhengxi and Yao, Zhiyuan and Han, Zhuowen and Wang, Zi-Han and Wu, Jinyang and Gu, Qi and Cai, Xunliang and Lu, Weiming and Xiao, Jun and Zhuang, Yueting and others},
  journal={arXiv preprint arXiv:2605.15155},
  year={2026}
}

@article{skillx,
  title={SkillX: Automatically constructing skill knowledge bases for agents},
  author={Wang, Chenxi and Yu, Zhuoyun and Xie, Xin and Yao, Wuguannan and Fang, Runnan and Qiao, Shuofei and Cao, Kexin and Zheng, Guozhou and Qi, Xiang and Zhang, Peng and others},
  journal={arXiv preprint arXiv:2604.04804},
  year={2026}
}

@article{skill2structure,
  title={From Skill Text to Skill Structure: The Scheduling-Structural-Logical Representation for Agent Skills},
  author={Liang, Qiliang and Wang, Hansi and Liang, Zhong and Liu, Yang},
  journal={arXiv preprint arXiv:2604.24026},
  year={2026}
}

@article{skillacq,
  title={Automating Skill Acquisition through Large-Scale Mining of Open-Source Agentic Repositories: A Framework for Multi-Agent Procedural Knowledge Extraction},
  author={Bi, Shuzhen and Wu, Mengsong and Hao, Hao and Li, Keqian and Liu, Wentao and Song, Siyu and Zhao, Hongbo and Zhou, Aimin},
  journal={arXiv preprint arXiv:2603.11808},
  year={2026}
}

@article{grpo,
  title={Deepseekmath: Pushing the limits of mathematical reasoning in open language models},
  author={Shao, Zhihong and Wang, Peiyi and Zhu, Qihao and Xu, Runxin and Song, Junxiao and Bi, Xiao and Zhang, Haowei and Zhang, Mingchuan and Li, YK and Wu, Yang and others},
  journal={arXiv preprint arXiv:2402.03300},
  year={2024}
}

@article{gigpo,
  title={Group-in-group policy optimization for llm agent training},
  author={Feng, Lang and Xue, Zhenghai and Liu, Tingcong and An, Bo},
  journal={Advances in Neural Information Processing Systems},
  volume={38},
  pages={46375--46408},
  year={2026}
}

@article{ragen,
  title={Ragen: Understanding self-evolution in llm agents via multi-turn reinforcement learning},
  author={Wang, Zihan and Wang, Kangrui and Wang, Qineng and Zhang, Pingyue and Li, Linjie and Yang, Zhengyuan and Jin, Xing and Yu, Kefan and Nguyen, Minh Nhat and Liu, Licheng and others},
  journal={arXiv preprint arXiv:2504.20073},
  year={2025}
}

@article{ST-PPO,
  title={ST-PPO: Stabilized Off-Policy Proximal Policy Optimization for Multi-Turn Agents Training},
  author={Li, Chenliang and Elmahdy, Adel and Boyd, Alex and Wang, Zhongruo and Garcia, Alfredo and Bhatia, Parminder and Kass-Hout, Taha and Xiao, Cao and Hong, Mingyi},
  journal={arXiv preprint arXiv:2511.20718},
  year={2025}
}

@inproceedings{collapse,
  title={Understanding Reasoning Collapse in Multi-Turn Agent Reinforcement Learning},
  author={Wang, Zihan and Gui, Chi and Jin, Xing and Wang, Qineng and Liu, Licheng and Wang, Kangrui and Chen, Shiqi and Li, Linjie and Yang, Zhengyuan and Zhang, Pingyue and others},
  booktitle={ICLR 2026 Workshop on Lifelong Agents: Learning, Aligning, Evolving},
  year={2026}
}

@article{unveiling,
  title={Unveiling Implicit Advantage Symmetry: Why GRPO Struggles with Exploration and Difficulty Adaptation},
  author={Yu, Zhiqi and Chen, Zhangquan and Liu, Mengting and Zhang, Heye and Qu, Liangqiong},
  journal={arXiv preprint arXiv:2602.05548},
  year={2026}
}

@article{too_correct,
  title={Too Correct to Learn: Reinforcement Learning on Saturated Reasoning Data},
  author={Liang, Zhenwen and Zhou, Yujun and Lu, Sidi and Zhang, Xiangliang and Mi, Haitao and Yu, Dong},
  journal={arXiv preprint arXiv:2604.18493},
  year={2026}
}

@article{turn-level,
  title={Reinforcing multi-turn reasoning in llm agents via turn-level reward design},
  author={Wei, Quan and Zeng, Siliang and Li, Chenliang and Brown, William and Frunza, Oana and Deng, Wei and Schneider, Anderson and Nevmyvaka, Yuriy and Zhao, Yang Katie and Garcia, Alfredo and others},
  journal={arXiv preprint arXiv:2505.11821},
  year={2025}
}

@article{spa-rl,
  title={Spa-rl: Reinforcing llm agents via stepwise progress attribution},
  author={Wang, Hanlin and Leong, Chak Tou and Wang, Jiashuo and Wang, Jian and Li, Wenjie},
  journal={arXiv preprint arXiv:2505.20732},
  year={2025}
}

@article{archer,
  title={Archer: Training language model agents via hierarchical multi-turn rl, 2024},
  author={Zhou, Yifei and Zanette, Andrea and Pan, Jiayi and Levine, Sergey and Kumar, Aviral},
  journal={URL https://arxiv. org/abs/2402.19446},
  year={2024}
}

@article{harnessing,
  title={Harnessing uncertainty: Entropy-modulated policy gradients for long-horizon llm agents},
  author={Wang, Jiawei and Liu, Jiacai and Fu, Yuqian and Li, Yingru and Wang, Xintao and Lin, Yuan and Yue, Yu and Zhang, Lin and Wang, Yang and Wang, Ke},
  journal={arXiv preprint arXiv:2509.09265},
  year={2025}
}

@article{drmas,
  title={Dr. MAS: Stable Reinforcement Learning for Multi-Agent LLM Systems},
  author={Feng, Lang and Zheng, Longtao and He, Shuo and Zhang, Fuxiang and An, Bo},
  journal={arXiv preprint arXiv:2602.08847},
  year={2026}
}

@article{coevoskills,
  title={Coevoskills: Self-evolving agent skills via co-evolutionary verification},
  author={Zhang, Hanrong and Fan, Shicheng and Zou, Henry Peng and Chen, Yankai and Wang, Zhenting and Zhou, Jiayu and Li, Chengze and Huang, Wei-Chieh and Yao, Yifei and Zheng, Kening and others},
  journal={arXiv preprint arXiv:2604.01687},
  year={2026}
}

@article{cognitive,
  title={Cognitive load during problem solving: Effects on learning},
  author={Sweller, John},
  journal={Cognitive science},
  volume={12},
  number={2},
  pages={257--285},
  year={1988},
  publisher={Elsevier}
}

@article{hdpo,
  title={Hdpo: Hybrid distillation policy optimization via privileged self-distillation},
  author={Ding, Ken},
  journal={arXiv preprint arXiv:2603.23871},
  year={2026}
}

@inproceedings{shortcut,
  title={Mitigating Shortcut Learning via Smart Data Augmentation based on Large Language Model},
  author={Sun, Xinyi and Tan, Hongye and Guo, Yaxin and Qiang, Pengpeng and Li, Ru and Zhang, Hu},
  booktitle={Proceedings of the 31st International Conference on Computational Linguistics},
  pages={8160--8172},
  year={2025}
}

@article{alfworld,
  title={Alfworld: Aligning text and embodied environments for interactive learning},
  author={Shridhar, Mohit and Yuan, Xingdi and C{\^o}t{\'e}, Marc-Alexandre and Bisk, Yonatan and Trischler, Adam and Hausknecht, Matthew},
  journal={arXiv preprint arXiv:2010.03768},
  year={2020}
}

@article{webshop,
  title={Webshop: Towards scalable real-world web interaction with grounded language agents},
  author={Yao, Shunyu and Chen, Howard and Yang, John and Narasimhan, Karthik},
  journal={Advances in Neural Information Processing Systems},
  volume={35},
  pages={20744--20757},
  year={2022}
}

@article{react,
  title={React: Synergizing reasoning and acting in language models},
  author={Yao, Shunyu and Zhao, Jeffrey and Yu, Dian and Du, Nan and Shafran, Izhak and Narasimhan, Karthik and Cao, Yuan},
  journal={arXiv preprint arXiv:2210.03629},
  year={2022}
}

@article{reflexion,
  title={Reflexion: Language agents with verbal reinforcement learning},
  author={Shinn, Noah and Cassano, Federico and Gopinath, Ashwin and Narasimhan, Karthik and Yao, Shunyu},
  journal={Advances in neural information processing systems},
  volume={36},
  pages={8634--8652},
  year={2023}
}

@article{mem0,
  title={Mem0: Building production-ready ai agents with scalable long-term memory},
  author={Chhikara, Prateek and Khant, Dev and Aryan, Saket and Singh, Taranjeet and Yadav, Deshraj},
  journal={arXiv preprint arXiv:2504.19413},
  year={2025}
}

@inproceedings{expel,
  title={Expel: Llm agents are experiential learners},
  author={Zhao, Andrew and Huang, Daniel and Xu, Quentin and Lin, Matthieu and Liu, Yong-Jin and Huang, Gao},
  booktitle={Proceedings of the AAAI Conference on Artificial Intelligence},
  volume={38},
  number={17},
  pages={19632--19642},
  year={2024}
}

@article{memp,
  title={Memp: Exploring agent procedural memory},
  author={Fang, Runnan and Liang, Yuan and Wang, Xiaobin and Wu, Jialong and Qiao, Shuofei and Xie, Pengjun and Huang, Fei and Chen, Huajun and Zhang, Ningyu},
  journal={arXiv preprint arXiv:2508.06433},
  year={2025}
}

@article{simplemem,
  title={SimpleMem: Efficient Lifelong Memory for LLM Agents},
  author={Liu, Jiaqi and Su, Yaofeng and Xia, Peng and Han, Siwei and Zheng, Zeyu and Xie, Cihang and Ding, Mingyu and Yao, Huaxiu},
  journal={arXiv preprint arXiv:2601.02553},
  year={2026}
}

@inproceedings{rloo,
  title={Back to basics: Revisiting REINFORCE-style optimization for learning from human feedback in LLMs},
  author={Ahmadian, Arash and Cremer, Chris and Gall{\'e}, Matthias and Fadaee, Marzieh and Kreutzer, Julia and Pietquin, Olivier and {\"U}st{\"u}n, Ahmet and Hooker, Sara},
  booktitle={Proceedings of the 62nd Annual Meeting of the Association for Computational Linguistics (Volume 1: Long Papers)},
  pages={12248--12267},
  year={2024}
}

@article{memrl,
  title={Memrl: Self-evolving agents via runtime reinforcement learning on episodic memory},
  author={Zhang, Shengtao and Wang, Jiaqian and Zhou, Ruiwen and Liao, Junwei and Feng, Yuchen and Li, Zhuo and Zheng, Yujie and Zhang, Weinan and Wen, Ying and Li, Zhiyu and others},
  journal={arXiv preprint arXiv:2601.03192},
  year={2026}
}

@article{evolver,
  title={Evolver: Self-evolving llm agents through an experience-driven lifecycle},
  author={Wu, Rong and Wang, Xiaoman and Mei, Jianbiao and Cai, Pinlong and Fu, Daocheng and Yang, Cheng and Wen, Licheng and Yang, Xuemeng and Shen, Yufan and Wang, Yuxin and others},
  journal={arXiv preprint arXiv:2510.16079},
  year={2025}
}

\appendix

\section{Related Work}

\subsection{Skill-Augmented Agentic Training}

Early work on agent skills predominantly employed skills as training-free in-context augmentations~\cite{skillx,skillacq,skill2structure,coevoskills}. However, constrained by model capacities and the increasing complexity of agentic tasks, recent research has expanded into training agents to effectively harness skills, which primarily diverge into two paradigms. One paradigm advocates for full externalization~\cite{skill1,metaclaw,arise,skillgraph}. SkillRL~\cite{skillrl}, for example, constructs a hierarchical skill bank where general skills are appended with task-specific skills, persistently maintaining all guidance within the contextual window throughout training and inference. Another line of research explores \textit{full internalization} to eliminate runtime context overhead~\cite{skillsd,sdar,hintsd}. SKILL0~\cite{skill0}, for instance, leverages a dynamic curriculum to progressively withdraw skills from the context until the agent operates without any external input, completely assimilating all guidance utility into model parameters.

Yet, joint skill internalization and utilization remains underexplored. A closely related work, SLIM~\cite{slim}, dynamically decides whether to retain skills for active utilization or retire them upon internalization. Still, SLIM treats all skills uniformly, and its final active skill set risks incompatibility with OOD tasks paired with unseen specific skills. In contrast, we explicitly decouple general and task-specific skills, jointly optimizing them for foundational internalization and adaptive utilization, respectively. This parameterizes foundational reasoning logic to actively exploit tailored external guidance in authentic OOD settings.

\subsection{Agentic Reinforcement Learning}
Reinforcement learning, particularly Group Relative Policy Optimization (GRPO)~\cite{grpo}, has become a core backbone for training LLMs as interactive agents. Recent algorithmic advances build upon this framework to address central challenges in multi-turn environments, including temporal credit assignment~\cite{gigpo,turn-level,spa-rl}, long-horizon optimization~\cite{archer,ST-PPO,harnessing}, and training stability against degenerate action cycles~\cite{ragen,drmas,collapse}. While these advances provide a solid optimization foundation, the objective of GRPO remains difficulty-agnostic---the intra-group reward variance collapses to zero on either impossibly hard tasks or overly simple tasks, which leads to exploration stagnation~\cite{unveiling} and saturation-induced mode collapse~\cite{too_correct}, respectively. Therefore, we dynamically perceive task difficulty and assign tailored auxiliary optimization objectives specifically for excessively hard and near-saturated tasks, ensuring the effectiveness of agentic RL training.

\section{WebShop Domain Split Statistics.}
\label{sec:split}
We use the 12,087 human-annotated goals from WebShop and partition them into seven domains via keyword-based classification of goal instructions. Four domains serve as ID categories for training and ID evaluation, while three are held out as out-of-distribution OOD categories for OOD evaluation only. To address the severe imbalance in the miscellaneous \textit{Other} category (which accounts for over 60\% of all goals before processing), we apply farthest-point sampling (FPS) based on sentence embeddings to downsample it to a scale comparable with the other domains.

The resulting per-domain statistics are as follows.
\textbf{Training set (ID, 3{,}320 goals):} Apparel 776, Electronics 938, Footwear 606, Other 1{,}000 (downsampled via FPS from ${\sim}$6{,}600).
\textbf{ID validation set (454 goals):} Apparel 113, Electronics 152, Footwear 89, Other 100 (downsampled via FPS from ${\sim}$940).
\textbf{OOD validation set (207 goals):} Accessories 80, Beauty \& Health 54, Home Decor 73.

The training range corresponds to goal indices 1{,}500+ in the original WebShop ordering, while the validation pool merges the original test (0--499) and development (500--1{,}499) splits to ensure sufficient per-domain coverage.

\section{Pseudo Code}
\begin{algorithm}[h]
\caption{\ours: Joint Skill Internalization and Utilization}
\label{alg:skill05}
\small
\begin{algorithmic}[1]
\REQUIRE Policy $\pi_\theta$, General Skills $\mathcal{S}_G$, Specific Skills $\mathcal{S}_S$
\FOR{each training step $t$}
    \STATE Sample batch $\mathcal{B}_t \sim \mathcal{X}_{train}^{id}$. Initialize $\mathcal{L} \gets 0$.
    
    \vspace{0.2em}
    \STATE \textbf{\% Phase-1: Difficulty-Aware Routing}
    \FOR{each $x_i \in \mathcal{B}_t$}
        \STATE $c_t^{std} \gets \mathcal{K}_t(x_i)$. Sample $G$ rollouts $\tau^{(1)} \sim \pi_\theta(\cdot|h_t, c_t^{std})$.
        \STATE Evaluate empirical pass rate $p_i$.
    \ENDFOR
    \STATE Compute dynamic routing threshold $\eta_t$ (Eq. 2).
    
    \vspace{0.2em}
    \STATE \textbf{\% Phase-2: Tier-Tailored Optimization}
    \FOR{each $x_i \in \mathcal{B}_t$}
        \IF{$p_i == 0$}
            \STATE \textit{\% Hard Tier}
            \STATE Sample $\tau^{(2)}$ guided by $c_t^{priv} \gets \mathcal{S}_G \cup \mathcal{K}_t(x_i)$.
            \STATE $\mathcal{L} \gets \mathcal{L} + \mathcal{L}_{hard}$ via token-level JSD (Eq. 4).
        \ELSIF{$p_i \le \eta_t$}
            \STATE \textit{\% Medium Tier}
            \STATE $\mathcal{L} \gets \mathcal{L} + \mathcal{L}_{medium}$ via standard GRPO on $\tau^{(1)}$ (Eq. 5).
        \ELSE
            \STATE \textit{\% Easy Tier}
            \STATE Sample diagnostic $\tau_{diag}$ using $c_t^{none} \gets \emptyset$ to get $p_i^{none}$.
            \STATE $A_i^u \gets$ offset derived from utilization gain $p_i - p_i^{none}$ (Eq. 6).
            \STATE $\mathcal{L} \gets \mathcal{L} + \mathcal{L}_{easy}$ via GRPO using $\hat{A}_i^{(g)} = A_i^{(g)} + A_i^u$.
        \ENDIF
    \ENDFOR
    
    \vspace{0.2em}
    \STATE Update policy parameters $\theta$ using $\nabla_\theta \mathcal{L}$.
\ENDFOR
\end{algorithmic}
\end{algorithm}

\section{Implementation Details}
\label{sec:immplement}

\paragraph{Inference Protocol.} 
To ensure a fair comparison, we strictly control the prompt conditions at test time to match each skill-bases method's design principle and OOD setting constraints. Specifically, \textbf{SkillRL} receives both general skills and ID or OOD-specific skills retrieved from the skill set. \textbf{SKILL0} receives no skill context on ID tasks due to full internalization, and receives only OOD-specific skills on OOD tasks. \textbf{SLIM} receives its final trained active skill set (comprising a subset of general and ID-specific skills) for ID testing, whereas its ID-specific skills are replaced with OOD ones for OOD evaluation. Our \textbf{\ours} receives only ID or OOD-specific skills during inference , as general skills have been internalized.

\begin{table*}[t]
\centering
\footnotesize
\setlength{\tabcolsep}{4.2pt}
\renewcommand{\arraystretch}{0.9}

\resizebox{0.85\textwidth}{!}{%
\begin{tabular}{lccccc|cccc}
\toprule
\multirow{2}{*}{\textbf{Method}}
& \multicolumn{5}{c|}{\textbf{ID}}
& \multicolumn{4}{c}{\textbf{OOD}} \\
\cmidrule(lr){2-6}
\cmidrule(lr){7-10}
& \textbf{Apparel}
& \textbf{Elec.}
& \textbf{Footwear}
& \textbf{Other}
& \textbf{Avg.}
& \textbf{Access.}
& \textbf{Beauty}
& \textbf{Home}
& \textbf{Avg.} \\
\midrule

\multicolumn{10}{l}{\emph{Prompt-based Methods}} \\
Zero-shot
& 4.4 & 4.6 & 3.4 & 1.0 & 3.5
& 2.5 & 3.7 & 5.5 & 3.9 \\

Few-shot
& 14.2 & 15.1 & 13.5 & 24.0 & 16.5
& 18.8 & 29.6 & 5.5 & 16.9 \\

\midrule
\multicolumn{10}{l}{\emph{Prompt-based Agentic or Memory-based Methods}} \\
ReAct
& 12.4 & 11.1 & 4.5 & 12.0 & 10.4
& 11.2 & 24.1 & 2.7 & 11.6 \\

Reflexion
& 3.5 & 8.6 & 1.1 & 7.0 & 5.5
& 6.3 & 5.6 & 1.4 & 4.4 \\

Mem0
& 8.9 & 9.2 & 2.3 & 11.0 & 8.2
& 10.0 & 16.7 & 4.1 & 9.7 \\

ExpeL
& 6.2 & 12.5 & 9.0 & 21.0 & 12.1
& 12.5 & 25.9 & 8.2 & 14.5 \\

MemP
& 15.9 & 13.2 & 9.0 & 19.0 & 14.3
& 16.2 & 14.8 & 9.6 & 13.5 \\

SimpleMem
& 11.5 & 13.8 & 6.7 & 13.0 & 11.7
& 10.0 & 20.4 & 5.5 & 11.1 \\

\midrule
\multicolumn{10}{l}{\emph{RL-based Methods}} \\

RLOO
& 34.9 & 23.9 & \cellcolor{best}\textbf{41.5} & 32.1 & 31.1
& 31.4 & 46.5 & 22.5 & 32.9 \\

GRPO
& 35.1 & 22.6 & 39.0 & \cellcolor{second}\underline{49.5} & 33.6
& 27.7 & 47.3 & 25.9 & 32.3 \\

\midrule
\multicolumn{10}{l}{\emph{Memory-Augmented RL Methods}} \\

MemRL
& 22.2 & 15.2 & 25.0 & 48.3 & 26.2
& 13.2 & 17.9 & 27.8 & 19.6 \\

EvolveR
& 32.5 & 31.1 & 25.0 & 20.9 & 28.0
& 20.8 & 28.6 & 18.2 & 21.9 \\

Mem0+GRPO
& 36.5 & 22.0 & 40.6 & 23.1 & 29.5
& 10.2 & 32.1 & 30.2 & 23.0 \\

SimpleMem+GRPO
& 25.4 & 28.0 & 26.6 & 25.1 & 26.4
& 16.2 & 32.1 & 29.4 & 25.0 \\

\midrule
\multicolumn{10}{l}{\emph{Skill-Augmented RL Methods}} \\

SkillRL
& 36.0
& 34.2
& \cellcolor{second}\underline{41.4} 
& 49.3
& \cellcolor{second}\underline{38.3}
& 36.3
& \cellcolor{second}\underline{48.5} 
& 27.6 
& \cellcolor{second}\underline{36.7} \\

SKILL0
& \cellcolor{best}\textbf{39.2} 
& 33.0 
& 38.1 
& 37.9 
& 35.2
& \cellcolor{best}\textbf{42.1} 
& 38.6 
& 26.5 
& 35.4 \\

SLIM
& 31.9 & \cellcolor{second}\underline{36.8} & 31.5 & 33.0 & 33.7
& 35.0 & 29.6 & \cellcolor{best}\textbf{35.6} & 33.8 \\

\textbf{\ours}
& \cellcolor{second}\underline{39.1} 
& \cellcolor{best}\textbf{37.3} 
& 41.1 
& \cellcolor{best}\textbf{50.9} 
& \cellcolor{best}\textbf{40.4}
& \cellcolor{second}\underline{36.6} 
& \cellcolor{best}\textbf{54.2} 
& \cellcolor{second}\underline{31.4} 
& \cellcolor{best}\textbf{40.6} \\

\bottomrule
\end{tabular}%
}

\caption{
Performance comparison on WebShop under ID and OOD task settings.
\colorbox{best}{\textbf{Best}} and
\colorbox{second}{\underline{second-best}}
results in each column are highlighted.
}
\label{tab:webshop_results}
\end{table*}

\paragraph{Training and Implementation Details.}
We use Qwen2.5-7B-Instruct as the base model. For policy optimization, we employ GRPO as the backbone with a group size of $G=8$, a learning rate of $1 \times 10^{-6}$. Training is conducted on 4 H800 GPUs with a batch size of 16 tasks per iteration, spanning 120 steps for ALFWorld and 150 steps for WebShop. The maximum interaction horizon is set to 30 steps for ALFWorld and 15 steps for WebShop. Task-specific skills are retrieved via Qwen3-Embedding-0.6B with a retrieval capacity of $K=3$. A sliding window of size $W=5$ is maintained for routing threshold and utilization gain tracking. For privileged distillation, the token-level JSD optimization is performed over the top-64 tokens following~\citet{hdpo}.

\section{Case Study}

\label{sec:case_study}

\begin{figure*}[t]
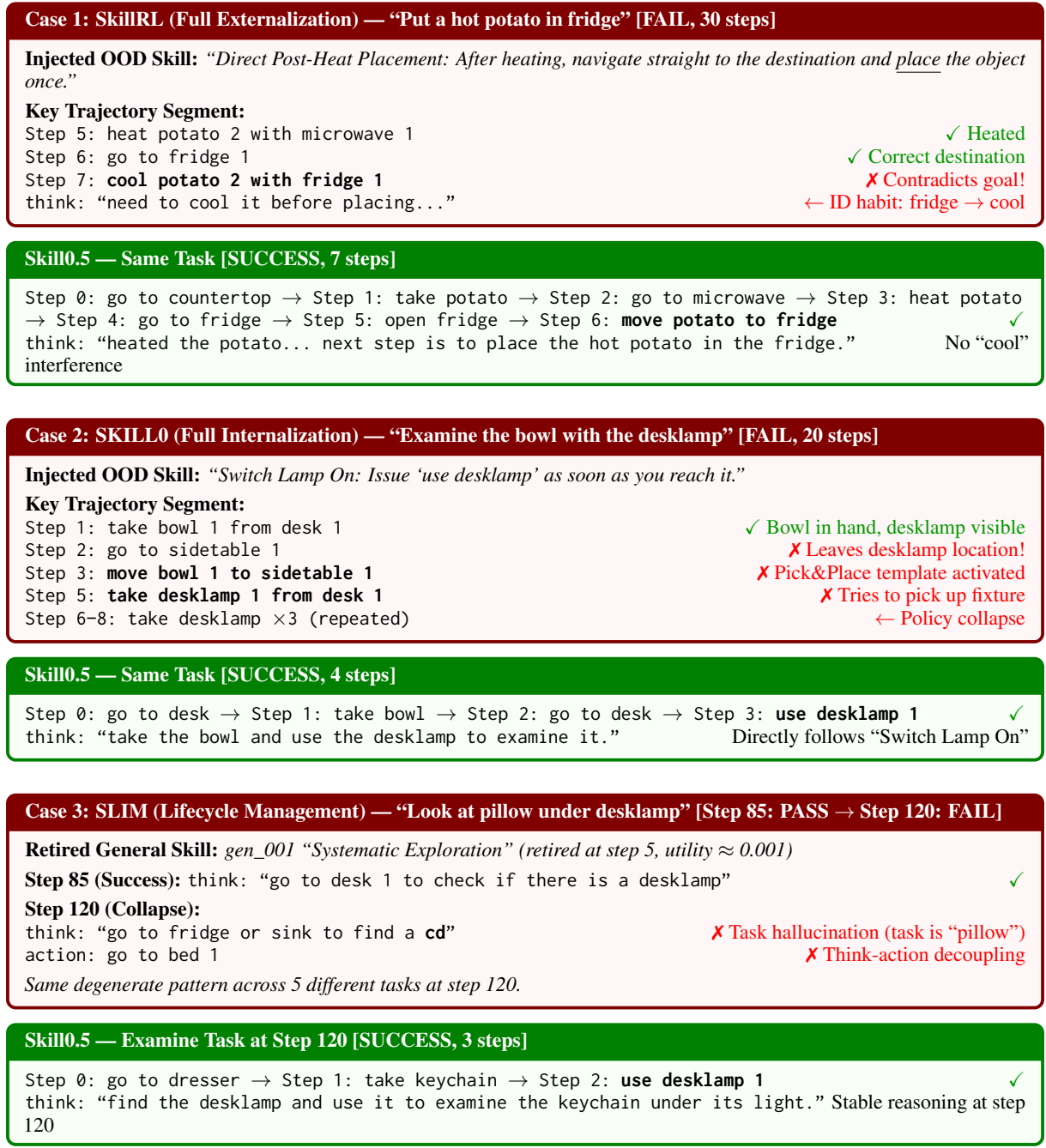

\centering
\footnotesize

\begin{tcolorbox}[
  colback=red!3, colframe=red!50!black,
  fonttitle=\bfseries\small,
  title={Case 1: SkillRL (Full Externalization) --- ``Put a hot potato in fridge'' [FAIL, 30 steps]},
  breakable, left=4pt, right=4pt, top=2pt, bottom=2pt,
  fontupper=\small\ttfamily
]
\normalfont\small
\textbf{Injected OOD Skill:} \textit{``Direct Post-Heat Placement: After heating, navigate straight to the destination and \underline{place} the object once.''}\\[3pt]
\textbf{Key Trajectory Segment:}\\
\texttt{Step 5: heat potato 2 with microwave 1} \hfill \textcolor{green!60!black}{\checkmark~Heated}\\
\texttt{Step 6: go to fridge 1} \hfill \textcolor{green!60!black}{\checkmark~Correct destination}\\
\texttt{Step 7: \textbf{cool potato 2 with fridge 1}} \hfill \textcolor{red}{\ding{55}~Contradicts goal!}\\
\texttt{~~think: ``need to cool it before placing...''} \hfill \textcolor{red}{$\leftarrow$ ID habit: fridge $\to$ cool}
\end{tcolorbox}

\vspace{-4pt}

\begin{tcolorbox}[
  colback=green!3, colframe=green!50!black,
  fonttitle=\bfseries\small,
  title={\ours{} --- Same Task [SUCCESS, 7 steps]},
  left=4pt, right=4pt, top=2pt, bottom=2pt,
  fontupper=\small\ttfamily
]
\normalfont\small
\texttt{Step 0: go to countertop $\to$ Step 1: take potato $\to$ Step 2: go to microwave $\to$ Step 3: heat potato}\\
\texttt{$\to$ Step 4: go to fridge $\to$ Step 5: open fridge $\to$ Step 6: \textbf{move potato to fridge}} \hfill \textcolor{green!60!black}{\checkmark}\\
\texttt{~~think: ``heated the potato... next step is to place the hot potato in the fridge.''} \hfill No ``cool'' interference
\end{tcolorbox}

\vspace{4pt}

\begin{tcolorbox}[
  colback=red!3, colframe=red!50!black,
  fonttitle=\bfseries\small,
  title={Case 2: SKILL0 (Full Internalization) --- ``Examine the bowl with the desklamp'' [FAIL, 20 steps]},
  breakable, left=4pt, right=4pt, top=2pt, bottom=2pt,
  fontupper=\small\ttfamily
]
\normalfont\small
\textbf{Injected OOD Skill:} \textit{``Switch Lamp On: Issue `use desklamp' as soon as you reach it.''}\\[3pt]
\textbf{Key Trajectory Segment:}\\
\texttt{Step 1: take bowl 1 from desk 1} \hfill \textcolor{green!60!black}{\checkmark~Bowl in hand, desklamp visible}\\
\texttt{Step 2: go to sidetable 1} \hfill \textcolor{red}{\ding{55}~Leaves desklamp location!}\\
\texttt{Step 3: \textbf{move bowl 1 to sidetable 1}} \hfill \textcolor{red}{\ding{55}~Pick\&Place template activated}\\
\texttt{Step 5: \textbf{take desklamp 1 from desk 1}} \hfill \textcolor{red}{\ding{55}~Tries to pick up fixture}\\
\texttt{Step 6--8: take desklamp $\times$3 (repeated)} \hfill \textcolor{red}{$\leftarrow$ Policy collapse}
\end{tcolorbox}

\vspace{-4pt}

\begin{tcolorbox}[
  colback=green!3, colframe=green!50!black,
  fonttitle=\bfseries\small,
  title={\ours{} --- Same Task [SUCCESS, 4 steps]},
  left=4pt, right=4pt, top=2pt, bottom=2pt,
  fontupper=\small\ttfamily
]
\normalfont\small
\texttt{Step 0: go to desk $\to$ Step 1: take bowl $\to$ Step 2: go to desk $\to$ Step 3: \textbf{use desklamp 1}} \hfill \textcolor{green!60!black}{\checkmark}\\
\texttt{~~think: ``take the bowl and use the desklamp to examine it.''} \hfill Directly follows ``Switch Lamp On''
\end{tcolorbox}

\vspace{4pt}

\begin{tcolorbox}[
  colback=red!3, colframe=red!50!black,
  fonttitle=\bfseries\small,
  title={Case 3: SLIM (Lifecycle Management) --- ``Look at pillow under desklamp'' [Step 85: PASS $\to$ Step 120: FAIL]},
  breakable, left=4pt, right=4pt, top=2pt, bottom=2pt,
  fontupper=\small\ttfamily
]
\normalfont\small
\textbf{Retired General Skill:} \textit{gen\_001 ``Systematic Exploration'' (retired at step 5, utility $\approx$ 0.001)}\\[3pt]
\textbf{Step 85 (Success):} \texttt{think: ``go to desk 1 to check if there is a desklamp''} \hfill \textcolor{green!60!black}{\checkmark}\\[3pt]
\textbf{Step 120 (Collapse):}\\
\texttt{think: ``go to fridge or sink to find a \textbf{cd}''} \hfill \textcolor{red}{\ding{55}~Task hallucination (task is ``pillow'')}\\
\texttt{action: go to bed 1} \hfill \textcolor{red}{\ding{55}~Think-action decoupling}\\[3pt]
\textit{Same degenerate pattern across 5 different tasks at step 120.}
\end{tcolorbox}

\vspace{-4pt}

\begin{tcolorbox}[
  colback=green!3, colframe=green!50!black,
  fonttitle=\bfseries\small,
  title={\ours{} --- Examine Task at Step 120 [SUCCESS, 3 steps]},
  left=4pt, right=4pt, top=2pt, bottom=2pt,
  fontupper=\small\ttfamily
]
\normalfont\small
\texttt{Step 0: go to dresser $\to$ Step 1: take keychain $\to$ Step 2: \textbf{use desklamp 1}} \hfill \textcolor{green!60!black}{\checkmark}\\
\texttt{~~think: ``find the desklamp and use it to examine the keychain under its light.''} \hfill Stable reasoning at step 120
\end{tcolorbox}

\caption{
Trajectory comparisons on ALFWorld OOD tasks. 
\ours\ succeeds in all cases by internalizing general reasoning while faithfully utilizing novel OOD-specific skills.
}
\label{fig:case_traj}
\end{figure*}

To qualitatively demonstrate why differentiated skill treatment is essential, we present representative OOD failure cases from each skill-augmented baseline on ALFWorld, contrasting them with \ours. Table~\ref{tab:case_summary} summarizes the failure mechanisms, and Figure~\ref{fig:case_traj} provides detailed trajectory comparisons.


\begin{table}[t]
\centering
\footnotesize
\setlength{\tabcolsep}{5pt}
\renewcommand{\arraystretch}{1.15}
\resizebox{\columnwidth}{!}{%
\begin{tabular}{@{} l l c c l @{}}
\toprule
\textbf{Method} & \textbf{OOD Task} & \textbf{Steps} & \textbf{Result} & \textbf{Failure Diagnosis} \\
\midrule
\rowcolor{red!8} SkillRL & Heat \& Place & 30 & \textcolor{red}{\ding{55}} & Contextual Interference \\
\rowcolor{green!6} \ours{} & Heat \& Place & 7 & \textcolor{green!60!black}{\checkmark} & --- \\
\midrule
\rowcolor{red!8} SKILL0 & Examine in Light & 20 & \textcolor{red}{\ding{55}} & Parametric Conflict \\
\rowcolor{green!6} \ours{} & Examine in Light & 4 & \textcolor{green!60!black}{\checkmark} & --- \\
\midrule
\rowcolor{red!10} SLIM & Examine in Light & 30 & \textcolor{red}{\ding{55}} & Behavioral Collapse \\
\rowcolor{green!6} \ours{} & Examine in Light & 3 & \textcolor{green!60!black}{\checkmark} & --- \\
\bottomrule
\end{tabular}%
}
\caption{Summary of representative OOD failure cases. Each baseline exhibits a distinct failure mechanism, while \ours\ consistently succeeds with significantly fewer steps.}
\label{tab:case_summary}
\end{table}


\paragraph{Case 1: Contextual Interference (SkillRL).}
In the ``Heat \& Place'' task, SkillRL's context window contains $\sim$1,617 tokens of general principles, common mistakes, and OOD-specific skills, with the latter occupying only $\sim$12\% of the total context. After successfully heating the potato and navigating to the fridge, the agent executes \texttt{cool potato with fridge}---directly contradicting both the task goal and the injected skill ``\textit{Direct Post-Heat Placement: place the object once}''. The agent's reasoning reveals that the ID-trained association ``fridge $\to$ cool'' (from Cool \& Place tasks) is activated by the general heuristic ``Use State-Changing Tools Early'', overpowering the novel OOD instruction. In contrast, \ours\ internalizes general skills into parameters, leaving only $\sim$214 tokens of OOD-specific guidance in the context. This ensures the novel ``place'' instruction receives undiluted attention, enabling correct execution in 7 steps.

\paragraph{Case 2: Parametric Knowledge Conflict (SKILL0).}
For the ``Examine in Light'' task, SKILL0 receives the OOD-specific skill ``\textit{Switch Lamp On: Issue `use desklamp'  as soon as you reach it}''. Despite the explicit instruction, the agent activates internalized ID procedural templates: it executes \texttt{move bowl to sidetable} (Pick \& Place terminal action) and repeatedly attempts \texttt{take desklamp} (treating it as a portable tool). Notably, the agent's reasoning even references the skill (``According to the Single Toggle Rule...'') yet immediately violates it---demonstrating that context-free training has atrophied the model's instruction-following capability, allowing parametric priors to dominate over novel textual guidance. \ours\ avoids this conflict entirely: general skills (domain-agnostic, non-procedural) are internalized, while task-specific skills remain external. The contrastive utilization training (\S\ref{sec:easy}) explicitly builds the ``read instruction $\to$ execute'' capability, enabling faithful compliance with novel OOD skills.

\paragraph{Case 3: Behavioral Collapse (SLIM).}
SLIM's lifecycle management retires ``Systematic Exploration'' at step 5 (utility $\approx$ 0.001) alongside 7 other general skills by step 50 (66.7\% retired). At step 85, the agent still succeeds on this task. However, by step 120---after 70 additional training steps without these general constraints---the policy exhibits catastrophic degradation: task hallucination (reasoning about ``cd'' when the task specifies ``pillow''), think-action decoupling (reasoning says ``fridge'' but action goes to ``bed''), and cross-task pattern collapse (identical degenerate template across 5 unrelated tasks). This irreversible cognitive erosion stems from treating general skills as retirable commodities rather than permanent cognitive foundations. \ours\ permanently embeds general skills via JSD distillation, making them immune to retirement decisions and ensuring stable reasoning throughout extended training.

\paragraph{Unified Diagnosis.}
These cases demonstrate that each uniform skill treatment strategy produces a characteristic failure mode in OOD settings: full externalization causes \textit{context-level} interference (fixable in principle by reducing context), full internalization causes \textit{parameter-level} conflict (unfixable at inference time), and indiscriminate lifecycle management causes \textit{temporal} degradation (irreversible once retired). \ours\ eliminates all three failure modes by design through its type-differentiated treatment: permanently internalizing domain-agnostic general skills while maintaining faithful utilization of dynamic task-specific skills.

\section{Detailed Hyperparameters}
\label{sec:hyper}
\begin{table}[h!]
\centering
\setlength{\tabcolsep}{6pt} 

\begin{tabular}{lc}
\toprule
\textbf{Hyperparameter} & \textbf{Value} \\
\midrule
Policy Optimization Backbone & GRPO \\
Learning Rate & 1e-6 \\
KL Regularization Coeff. & 0.01 \\
Entropy Bonus Coeff. & 0.001 \\
Invalid Action Penalty Coeff. & 0.1 \\
GRPO Group Size & 8 \\
Batch size & 16 \\
Mini-Batch size & 128 \\
Token-level Top-k for JSD Loss & 64 \\
Maximum Prompt Token Length & 6000 \\
Maximum Response Token Length & 768  \\
Evaluation Sampling Temperature & 0.4 \\
\bottomrule
\end{tabular}
\caption{Detailed configuration of training hyperparameters.}
\label{tab:hyperparameters}
\end{table}

\end{document}